\documentclass[10pt]{article}
\usepackage[margin=1in]{geometry}
\usepackage[colorlinks,citecolor=red]{hyperref}

\usepackage{graphicx}
\DeclareGraphicsExtensions{.pdf}
\usepackage[justification=Centering,singlelinecheck=false,caption=false,font=footnotesize]{subfig}

\usepackage{natbib}
\bibpunct{[}{]}{,}{n}{}{;}

\usepackage[cmex10]{amsmath}
\usepackage{bm}
\numberwithin{equation}{section}

\usepackage{amsmath}
\usepackage{amsfonts}
\usepackage{amssymb}
\usepackage{color}
\usepackage{amsthm}

\usepackage{url}
\usepackage{algorithm}
\usepackage{algorithmic}
\usepackage{subfig}
\usepackage{array}
\usepackage{amsthm}

\usepackage{multirow}
\usepackage{array}

\usepackage[ruled,linesnumbered,algo2e]{algorithm2e}
\usepackage{textcomp}
\usepackage{stfloats}

\begin{document}

\title{\textbf{Robust Learning under Hybrid Noise}}

\author{Tongliang Liu \hspace{1cm} Dacheng Tao \\
Centre for Quantum Computation \& Intelligent Systems\\Faculty of Engineering \& Information Technology\\University of Technology Sydney\\81-115 Broadway, Ultimo, NSW\\Australia\\
Email: tliang.liu@gmail.com, dacheng.tao@uts.edu.au}

\author{Yang Wei$^a$\hspace{0.5cm}Shuo Chen$^b$\hspace{0.5cm}Shanshan Ye$^c$\hspace{0.5cm}Bo Han$^d$\hspace{0.5cm}Chen Gong$^a$\\
School of Computer Science and Engineering, Nanjing University of Science and Technology, China\\
Center for Advanced Intelligence Project (AIP),
RIKEN National Science Institute, Japan\\
Australian Artificial Intelligence Institute, Faculty of Engineering and IT, Sydney\\
Department of Computer Science, Faculty of Science, Hong Kong Baptist University,  
}

\date{04 July 2024}

\maketitle

\begin{abstract}
 Feature noise and label noise are ubiquitous in practical scenarios, which pose great challenges for training a robust machine learning model. Most previous approaches usually deal with only a single problem of either feature noise or label noise. However, in real-world applications, hybrid noise, which contains both feature noise and label noise, is very common due to the unreliable data collection and annotation processes. Although some results have been achieved by a few representation learning based attempts, this issue is still far from being addressed with promising performance and guaranteed theoretical analyses. To address the challenge, we propose a novel unified learning framework called ``\textbf{F}eature and \textbf{L}abel \textbf{R}ecovery''~(FLR) to combat the hybrid noise from the perspective of data recovery, where we concurrently reconstruct both the feature matrix and the label matrix of input data. Specifically, the clean feature matrix is discovered by the low-rank approximation, and the ground-truth label matrix is embedded based on the recovered features with a nuclear norm regularization. Meanwhile, the feature noise and label noise are characterized by their respective adaptive matrix norms to satisfy the corresponding maximum likelihood. As this framework leads to a non-convex optimization problem, we develop the non-convex Alternating Direction Method of Multipliers~(ADMM) with the convergence guarantee to solve our learning objective. We also provide the theoretical analysis to show that the generalization error of FLR can be upper-bounded in the presence of hybrid noise. Experimental results on several typical benchmark datasets clearly demonstrate the superiority of our proposed method over the state-of-the-art robust learning approaches for various noises.
\end{abstract}


\section{Introduction}
The successes of most existing machine learning algorithms usually depend on the availability of high-quality data with clean features and accurate annotated labels. However, such high-quality data can hardly be guaranteed in lots of real-world scenarios due to various subjective and objective factors such as the insufficient expert knowledge~\cite{gong2017learning} and uncontrollable measurement error~\cite{grubbs1973errors}. 

Although many previous methods have designed robust classifiers to solely tackle the feature noise or label noise, they are still in the mess of performance degradation when the hybrid noise occurs in training data. Hybrid noise is usually induced by simultaneously corrupted features and labels, and it is very common in many real-world applications. For instance, if we have numerous image examples that are collected from visual sensors and are annotated manually, those images may be contaminated by non-ideal illumination or occlusion~\cite{chen2017low}; also, labeling errors are inevitably introduced due to human fatigues~\cite{Magoulas2001,tsouvalas2024labeling}. In this case, the hybrid noise will significantly damage the model training and severely decline the classification accuracy because the input data and output prediction are both misled by the feature noise and label noise, respectively. Therefore, a new unified learning algorithm is desired to deal with the hybrid noise in practical applications. 

\begin{figure}[t]
\centering
\includegraphics[width=0.75\columnwidth]{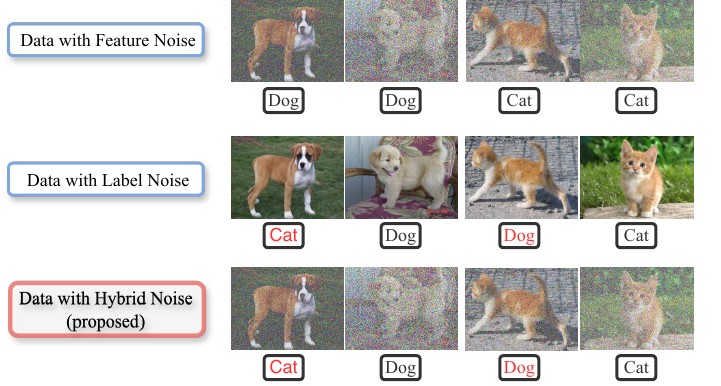} 
\caption{The problem setting of our proposed hybrid noise learning. Data with Feature Noise: all labels are correct, yet the features of examples are corrupted. Data with Label Noise: the features of all examples are clean, while some labels are incorrect. We consider a more challenging case, namely Data with Hybrid Noise: both features and labels of training examples are noisy. The noisy label is indicated in red.}
\label{fig: the setting of hybrid noise}
\end{figure}

Most existing methods for tackling label noise mainly focus on picking up clean labeled examples from the raw training data, such as MentorNet~\cite{jiang2018mentornet} and Co-teaching~\cite{han2018co}. Yet these kinds of works cannot theoretically guarantee the label correctness of the selected clean examples, so they can hardly obtain stable performance in practical uses. Therefore, some previous works, including $\mu$ Stochastic Gradient Descent~($\mu$SGD)~\cite{patrini2016lossFactorizationWeaklySupervisedLearning} and Symmetric Cross Entropy~(SCE)~\cite{wang2019symmetric}, further consider directly designing the different robust loss functions, avoiding the above sample selection. Nevertheless, they inevitably become weak when learning with some complicated noise, because they do not explicitly characterize the generation process of the label noise. In addition, there are some other approaches which propose to correct the data distribution based on the estimated statistics, such as transition matrix~\cite{han2018masking} and dataset centroid~\cite{gong2019loss}. However, all the above methods share an implicit deficiency, where they assume the features of all examples are completely clean. It means that these label-noise-specific learning algorithms cannot effectively handle the hybrid noise.

\begin{figure}[t]
\centering
\includegraphics[width=0.55\columnwidth]{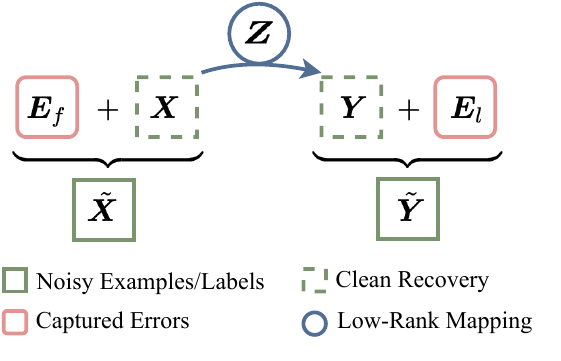} 
\caption{The proposed unified learning paradigm for the hybrid noise removal. The clean feature matrix $\bm{X}$ and correct label matrix $\bm{Y}$ are recovered from noisy data $\tilde{\bm{X}}$, $\tilde{\bm{Y}}$, respectively. Meanwhile, the true label matrix $\bm{Y}$ is embedded by $\bm{X}$ via a low-rank projection $\bm{Z}$, where the feature error matrix is $\bm{E}_f$ and the label error matrix is $\bm{E}_l$, respectively.}
\label{fig: the motivation of proposed method}
\end{figure}

On the other hand, dealing with feature noise is also a classical topic in the robust learning area. To be specific, reconstruction based methods and regression based methods have shown the great effectiveness in some common noise cases, including the Gaussian noise and Laplacian noise~\cite{gao2008robust}. For example, Robust Principal Component Analysis~(RPCA)~\cite{candes2011robust} and Low-Rank Representation~(LRR)~\cite{liu2010robust} employ $\ell_1$-norm or $\ell_{2,1}$-norm to characterize the reconstruction residual which follows the Laplacian distribution. Moreover, the Frobenius norm $\Vert \cdot \Vert_F$ can be utilized to characterize the Gaussian distributed noise~\cite{chen2013robust} in some popular models such as denoising Autoencoder~\cite{vincent2010stacked}. Note that all the above reconstruction based algorithms usually need an additional classifier to perform classification tasks. In contrast, some regression based methods including Sparse Representation Classifier~(SRC)~\cite{wright2008robust} and Nuclear-Norm based Matrix Regression~(NMR)~\cite{yang2016nuclear} propose to directly classify examples based on the reconstruction error, but they require fully correct labels as supervision, so they are still unable to handle the hybrid noise. Note that although Robust Representation Learning~(RRL)~\cite{li2021learning} can address label noise, out-of-distribution input, and input corruption simultaneously, it is still short of the explicit module to handle feature noise and the theoretical guarantee of denoising capability. In addition, RRL is limited to dealing with noisy data in visual tasks, while hybrid noise is also widespread in various non-visual real-world tasks.

To overcome the above drawbacks of existing robust learning approaches, we propose a novel unified learning paradigm called ``\textbf{F}eature and \textbf{L}abel \textbf{R}ecovery''~(FLR), which allows us to train a robust classifier with the heavy hybrid noise. The setting of hybrid noise is shown in Figure~\ref{fig: the setting of hybrid noise}, which describes the main difference among the hybrid noise, label noise, and feature noise. Specifically, the red box shows the hybrid noise case where four corrupted animal images belong to two classes~(here the first and third examples are mislabeled). To address this challenging setting, we construct a new data recovery framework to simultaneously learn both the clean feature matrix and the true label matrix in a single optimization objective. Given the observed noisy feature matrix $\tilde{\bm{X}}$ and the corresponding noisy label matrix $\tilde{\bm{Y}}$, the clean feature matrix $\bm{X}$ is a low-rank approximation to its noisy matrix $\tilde{\bm{X}}$~(as shown in Figure~\ref{fig: the motivation of proposed method}), and the feature error $\bm{E}_f$ is characterized by adaptive matrix norms under different types of noise assumption~(\emph{i.e.}, $\tilde{\bm{X}} = \bm{X} + \bm{E}_f$) to satisfy the corresponding maximum likelihood. Furthermore, the correct label matrix $\bm{Y}$ is embedded based on the recovered feature matrix $\bm{X}$ with a low-rank projection $\bm{Z}$~(\emph{i.e.}, $\bm{Y} = \bm{XZ}$), so that the label noise is captured by a row-sparse matrix $\bm{E}_l$. 

FLR seamlessly integrates the hybrid noise removal and classifier learning~(as the projection matrix $\bm{Z}$ predicts class labels for all examples) into a unified optimization model, which is solved by our designed non-convex Alternating Direction Method of Multipliers~(ADMM). We prove that our iteration algorithm can converge to a stationary point of the learning objective. Meanwhile, we also provide theoretical analyses to reveal that both the clean feature matrix and the true label matrix can be correctly recovered with increasing example size. Thanks to the integrated consideration of the feature noise and label noise, as well as the effectiveness of the low-rank matrix recovery technique, our proposed method achieves better results than existing robust learning approaches. The main contributions of our paper are summarized below:
\begin{itemize}
    \item We propose the hybrid noise problem formally~(\emph{i.e.}, the integration of both the feature noise and label noise), and we mathematically define this practical problem setting to a general data recovery objective.
    \item  We propose a unified learning framework to deal with the hybrid noise problem, where we provide comprehensive theoretical analyses to guarantee the algorithm convergence as well as the generalization ability of our method.
    \item We conduct intensive experiments on popular benchmark datasets to demonstrate the superiority of our method over the state-of-the-art robust learning approaches.
\end{itemize}

\section{Related Work}
In this section, we briefly review representative works on label noise learning and feature noise learning.

\subsection{Label Noise Learning}
The main goal of label noise learning is to train a robust classifier with noisy supervision to avoid the degradation of classification performance on test data.

A straightforward idea is to improve the label reliabilities of a small part of examples. The early-stage approaches~\cite{muhlenbach2004identifying, miranda2009use} firstly identify correctly labeled examples, and then learn the corresponding classifier with selected clean examples. Recently, the deep neural network based method MentorNet~\cite{jiang2018mentornet} learns a data-driven curriculum to pick up the examples with potentially correct labels. Co-teaching~\cite{han2018co} simultaneously trains two networks in a cooperative manner, where each network selects the small-loss examples for its peer network training. DivideMix~\cite{li2019dividemix} discards noisy labels but preserves the corresponding instances, and then trains DNNs in the manner of semi-supervised learning. RTLC~\cite{zhang2023noise} and CLC~\cite{zeng2022clc} aim to identify the noisy labels and correct them in Federated Learning. Nevertheless, it is hard to verify the label correctness of the selected examples, which makes these kinds of methods not always reliable in practical uses.

To avoid the unreliable process of sample selection, LICS~\cite{gao2016risk} and $\mu$SGD~\cite{patrini2016lossFactorizationWeaklySupervisedLearning} decompose the loss functions into a label-independent term and a label-dependent term, and then only correct the second term to reduce the negative effect caused by noisy labels. Ghosh et al.~\cite{ghosh2017robust} further generalize the existing noise-tolerant loss functions to multi-class classification problems.

Furthermore, some methods consider tackling the label noise based on the statistics estimation. Masking~\cite{han2018masking} conveys human cognition of invalid class transitions and naturally speculates the structure of the noise transition matrix. 
However, existing estimators, especially those anchor points or cluster based methods, generally require informative representations. Therefore, Zhu et al.~\cite{zhu2022beyond} build the transition matrix estimator using the distilled features, which can just handle the less informative features, but it does not take the feature noise into account.
\subsection{Feature Noise Learning}
The original idea for feature noise learning is to recover the latent clean examples from the corrupted data and feed the reconstructed examples to downstream tasks. Both RPCA~\cite{candes2011robust} and LRR~\cite{liu2010robust} aim to reconstruct corrupted examples with the low-rank constraint, where RPCA assumes that the clean data matrix is low-rank while LRR assumes the representation coefficient matrix is low-rank. Chen et al.~\cite{chen2019delta} further propose the $\delta$-norm to characterize the structural noise, so that the error matrix is structurally sparse during the training phase. Recently, several deep neural network based methods have been proposed to handle noisy examples. CFMNet~\cite{du2024flexible} conducts the image noise removal with multi-layer conditional feature modulations. NIM-AdvDef~\cite{you2024beyond} adopts denoising as the pre-text task, and reconstructs noisy images well despite severe corruption. 

Note that all the above reconstruction based algorithms usually need an additional classifier to perform classification tasks. In contrast, some regression based methods such as SRC~\cite{wright2008robust} and NMR~\cite{yang2016nuclear} propose to directly classify examples based on the minimum reconstruction strategy. However, all methods mentioned above still require fully correct labels as supervision. As a result, these approaches are unable to handle hybrid noise. RRL~\cite{li2021learning} can handle out-of-distribution input and input corruption in visual tasks, but no specific module is designed for tackling feature noise. This motivates us to propose a completely new learning framework to simultaneously recover clean data from noisy features and noisy labels~(\emph{i.e.}, the hybrid noise).

\section{Our Proposed FLR}
In this section, we firstly describe our new method FLR and then provide the optimization algorithm to solve it.

\subsection{Model Formulation} 
We assume that the data matrix $\tilde{\bm{X}}\in \mathbb{R}^{n\times d}$ is partially corrupted by noise $\bm{E}_f \in \mathbb{R}^{n \times d}$, and the latent clean training feature matrix $\bm{X} = \bm{\tilde{X}} - \bm{E}_f$, where $n$ is the number of training examples, and $d$ indicates the feature dimension. $\bm{\tilde{Y}} \in  \mathbb{R}^{n \times c}$ is the matrix of corresponding observed noisy labels~($c$ denotes the number of classes). The element $\tilde{\bm{Y}}_{i,j} = 1$ if the $i$-th example is labeled as $j$, and $\tilde{\bm{Y}}_{i,j} = 0$ otherwise, where $i = 1,2, \dots, n$ and $j = 1,2, \dots, c$. 

\subsubsection{Low-Rank Projection from Feature to Label}

To be more specific, here the noisy $\bm{\tilde{Y}}$ is also decomposed into two parts, \emph{i.e.}, $\bm{\tilde{Y}} = \bm{Y} +\bm{E}_l$. The first term is the correct label matrix $\bm{Y} = \bm{X} \bm{Z}$, which is embedded based on the recovered feature matrix $\bm{X}$ with the projection matrix $\bm{Z}$. The second term $\bm{E}_l$ describes the error between the observed label and the accurate label. Moreover, it is reasonable to assume that the clean label matrix $\bm{Y} = \bm{X} \bm{Z}$ lies in a low-dimensional subspace~\cite{xu2016robust}, as it is one-hot encoded with only $c$ different classes. Therefore, we encourage the ground-truth label matrix $\bm{Y} = \bm{X} \bm{Z}$ to be low-rank by constraining each factor matrix.

To be more specific, due to the high intra-class similarities of features, the latent clean feature matrix $\bm{X}$ is inherently low-rank, and it can be recovered from the two equality constraints regarding the feature matrix and label matrix. As a matter of course, the projection matrix $\bm{Z}$ is also low-rank, because we only need a small number of columns in $\bm{X}$ to construct the low-rank label space $\bm{Y} = \bm{XZ}$. Finally, here $\bm{Y} = \bm{X} \bm{Z}$ is assumed as the true label matrix, so each element in it should be either 0 or 1. 

\subsubsection{Hybrid Noise Characterization}
Considering the complexity and diversity of the feature noise, we have to leverage different norm metrics~(\emph{e.g.}, $\Vert\bm{E}_f\Vert_F^2$, $\Vert\bm{E}_f\Vert_1$, {\emph{etc.}}) to constrain the feature noise matrix $\bm{E}_f$. Meanwhile, the number of the non-zero rows of label error matrix $\bm{E}_l$ is supposed to be small because those noisy labels are corrected while the remaining labels are still preserved. In this view, the label noise is characterized by the $\ell_{2,1}$-norm. By integrating the above regularization terms and contractions into a joint learning framework, we naturally have the following learning objective with double low-rank constraints:
\begin{equation}\label{eq: our proposed model for feature and label correction hard}
	\begin{aligned}
\setlength{\abovedisplayskip}{0pt}
\setlength{\belowdisplayskip}{0pt}
	&\min_{\substack{\bm{Z},\bm{X}, \\ \bm{E}_f, \bm{E}_l}}  \Vert  \bm{X}\Vert_*  + \lambda_1 \Vert\bm{Z}\Vert_*  +  \lambda_2 \mathrm{R}(\bm{E}_f)  +  \lambda_3\Vert\bm{E}_l\Vert_{2,1} \\
	&\text{s.t.}\ \bm{\tilde{X}}= \bm{X} + \bm{E}_f, \bm{\tilde{Y}} = \bm{X}\bm{Z} + \bm{E}_l, \bm{XZ}\in\left\{0,1\right\}^{n\times c},
   \end{aligned}
\end{equation}
where $\mathrm{R}(\bm{E}_f)$ is the generalized regularization form of the feature noise $\bm{E_f}$. Here $\lambda_1$, $\lambda_2$, $\lambda_3$ $> 0$ are trade-off parameters. We note that Eq.~\eqref{eq: our proposed model for feature and label correction hard} falls into an integer programming problem, which is generally NP-hard. To make Eq.~\eqref{eq: our proposed model for feature and label correction hard} tractable, we relax the discrete constraint to a continues condition $\bm{XZ} \in [0,1]^{n \times c}$, which is a linear relaxation and has been used in several prior works~\cite{hsieh2015pu, wei2019harnessing}. By solving the linear relaxation problem of Eq.~\eqref{eq: our proposed model for feature and label correction hard}, we can obtain the optimal projection matrix $\bm{Z}$, which can be subsequently applied as a projection based classifier. Since this classifier is learned from the denoising data, it is robust to hybrid noise, so that we can successfully predict the categories of the unseen test examples.

\subsection{Optimization}
\label{subsection: the optimization algorithm for the proposed method}
We develop a unified optimization framework for both hybrid noise removal and classier learning. Our complex problem setting leads to a non-convex objective function. In this section, the non-convex ADMM is applied to address our proposed model with the $\ell_1$-norm to constrain the feature noise $\bm{E}_f$.

By introducing three auxiliary variables and conducting linear relaxation, Eq.~\eqref{eq: our proposed model for feature and label correction hard} is transformed to the following equivalent form:
\begin{equation}\label{eq: equivalent form for feature and label correction with auxiliary variables}
\setlength{\abovedisplayskip}{0pt}
\setlength{\belowdisplayskip}{0pt}
\begin{aligned}
&\min_{\substack{\bm{Z},\bm{X},\bm{B},\bm{J},\\ \bm{K}, \bm{E}_f, \bm{E}_l}} \Vert \bm{X}\Vert_* +  \lambda_1\Vert\bm{Z}\Vert_*  +  \lambda_2\Vert \bm{E}_f\Vert_1  +  \lambda_3\Vert\bm{E}_l\Vert_{2,1} ,\\
& \ \ \ \ \ \text{s.t.}\ \bm{\tilde{X}}= \bm{X} + \bm{E}_f, \bm{\tilde{Y}} = \bm{B} + \bm{E}_l, \bm{Z}=\bm{J}, \\
& \ \ \ \ \ \ \ \ \ \ \bm{X} = \bm{K}, \bm{B}=\bm{KJ}, \bm{B}\in\left[0, 1\right]^{n\times c},
\end{aligned}
\end{equation}
and the augmented Lagrangian function of Eq.~\eqref{eq: equivalent form for feature and label correction with auxiliary variables} with the continuous convex constraint is considered as:
\begin{equation}
	\begin{aligned}
\setlength{\abovedisplayskip}{0pt}
\setlength{\belowdisplayskip}{0pt}
 \label{eq: the lagrangian function of feature and label noise}
	\mathcal{L}&\left(   \bm{X},  \bm{Z},  \bm{B},  \bm{J},  \bm{K},  \bm{E}_f,  \bm{E}_l,  \bm{M}_1,  \bm{M}_2,  \bm{M}_3,  \bm{M}_4,  \bm{M}_5  \right) \\
	& =  \Vert \bm{X}\Vert_* + 
	        \lambda_1 \Vert\bm{Z}\Vert_* + 
	        \lambda_2 \Vert \bm{E}_f\Vert_1 + 
	        \lambda_3 \Vert\bm{E}_l\Vert_{2,1} \\ 
	& + \mathrm{tr}(\bm{M}_1^\top(\bm{\tilde{X}} - \bm{X} - \bm{E}_f))
	    + \mathrm{tr}(\bm{M}_2^\top(\bm{\tilde{Y}} - \bm{B} - \bm{E}_l))\\
	& + \mathrm{tr}(\bm{M}_3^\top(\bm{Z} - \bm{J})) + 
			\mathrm{tr}(\bm{M}_4^\top(\bm{B} - \bm{KJ}))\\  
	& +\mathrm{tr}(\bm{M}_5^\top(\bm{X} - \bm{K})) 
	    +\frac{\mu}{2}\left(\Vert\bm{\tilde{X}} - \bm{X} - \bm{E}_f\Vert_F^2 +  \Vert\bm{Z}  -  \bm{J}\Vert_F^2 \right.\\
	& \left.+ \Vert\bm{\tilde{Y}}  -  \bm{B}  -  \bm{E}_l\Vert_F^2   +  \Vert\bm{B}  -  \bm{KJ}\Vert_F^2 + \Vert\bm{X} - \bm{K}\Vert_F^2\right),
	\end{aligned}
\end{equation}
where $\bm{M}_1$, $\bm{M}_2$, $\bm{M}_3$, $\bm{M}_4$, and $\bm{M}_5$ are Lagrangian multipliers, and $\mu>0$ is the penalty coefficient.

Under the framework of ADMM, we can alternately minimize each of the variables $\bm{Z},\bm{X},\bm{B},\bm{J},\bm{K}, \bm{E}_f$ and $\bm{E}_l$ by keeping the others fixed in each iteration.

\noindent
\textbf{Update $\bm{X}$:} The subproblem of $\bm{X}$ is
	\begin{equation}\label{eq: the subprblem of X}
\setlength{\abovedisplayskip}{0pt}
\setlength{\belowdisplayskip}{0pt}
	\begin{aligned}	
 &\min\limits_{\bm{X}}
	 	\Vert \bm{X}\Vert_* +                         
	 	\mathrm{tr}(\bm{M}_1^\top(\bm{\tilde{X}}-\bm{X}-\bm{E}_f)) + \mathrm{tr}(\bm{M}_5^\top(\bm{X}-\bm{K}))\\
	 	&\ \ \ \ \ + \frac{\mu}{2}(\Vert\bm{\tilde{X}}-\bm{X} -
	 	\bm{E}_f\Vert_F^2 +
 		\Vert\bm{X}-\bm{K}\Vert_F^2)\\
	\Rightarrow& 
 \min\limits_{\bm{X}} \tau\Vert\bm{X}\Vert_* + \frac{1}{2}\Vert\bm{X - \hat{\bm{X}}}\Vert_F^2,
	\end{aligned}
	\end{equation}
in which $\tau = \frac{1}{2\mu}$ and $\hat{\bm{X}} = \frac{\mu(\bm{\tilde{X}}-\bm{E}_f+\bm{K})-\bm{M}_5+\bm{M}_1}{2\mu}$.
The closed-form solution to Eq.~\eqref{eq: the subprblem of X} is 
\begin{equation}\label{eq: the solution to subproblem X}
	\begin{aligned}
	\bm{X} = \bm{U}_x   diag   \left(   \max \left\{   {\bm{\Sigma}_x}_{(ii)}-\tau, 0   \right\}   \right)\bm{V}_x^\top,
	\end{aligned}
\end{equation}
where $\forall i = 1, 2, \dots, \min(n, d)$, $\bm{U}_x$ and $\bm{V}_x$ are obtained by conducting the Singular Value Decomposition~(SVD) on $\hat{\bm{X}}$~({\em{i.e.}}, $\hat{\bm{X}} =  \bm{U}_x\bm{\Sigma}_x\bm{V}_x^\top$), and ${\bm{\Sigma}_x}_{(ii)}$ is the $i$-th diagonal element of the singular value matrix $\bm{\Sigma}_x$.

\noindent
\textbf{Update $\bm{Z}$:}
The subproblem related to the variable $\bm{Z}$ is
\begin{equation}
\begin{aligned}
	&\min\limits_{\bm{Z}} 
	\lambda_1\Vert\bm{Z}\Vert_* +
	\mathrm{tr}(\bm{M}_3^\top(\bm{Z} - \bm{J})) + 
	\frac{\mu}{2}\Vert\bm{Z} - \bm{J}\Vert_F^2\\
	\Rightarrow
	&\min\limits_{\bm{Z}} 
	\eta\Vert\bm{Z}\Vert_* +
	\frac{1}{2}\Vert\bm{Z} - \hat{\bm{Z}}\Vert_F^2,
\end{aligned}
\end{equation}
where $\eta = \frac{\lambda_1}{\mu}$ and $\hat{\bm{Z}} =  \bm{J} - \frac{\bm{M}_3}{\mu}$. Similarly to the closed-form solution Eq.~\eqref{eq: the solution to subproblem X}, we have
\begin{equation}\label{eq: the solution to subproblem Z}
\setlength{\abovedisplayskip}{0pt}
\setlength{\belowdisplayskip}{0pt}
	\begin{aligned}
	\bm{Z} = \bm{U}_z   diag   \left(   \max \left\{   {\bm{\Sigma}_z}_{(ii)}-\eta, 0   \right\}   \right)\bm{V}_z^\top,
	\end{aligned}
\end{equation}
in which $\forall i = 1, 2, \dots, \min(d, c)$.

\noindent
\textbf{Update $\bm{B}$:} The subproblem on $\bm{B}$ with the continuous convex set $\bm{B} \in [0,1]^{n \times c}$ is
\begin{equation} \label{eq: the subproblem of B}
\setlength{\abovedisplayskip}{0pt}
\setlength{\belowdisplayskip}{0pt}
	\begin{aligned}
		&\min\limits_{\bm{B}} 
		\mathrm{tr}(\bm{M}_2^\top(\bm{\tilde{Y}} - \bm{B} - \bm{E}_l)) +
		\mathrm{tr}(\bm{M}_4^\top(\bm{B} - \bm{KJ})) \\ 
		&\ \ \ \ \ + \frac{\mu}{2}\left(\Vert\bm{\tilde{Y}} - \bm{B} - \bm{E}_l\Vert_F^2 + \Vert\bm{B} - \bm{KJ}\Vert_F^2\right).
	\end{aligned}
\end{equation}
Obviously, the optimal $\bm{B}^\ast$ to Eq.~\eqref{eq: the subproblem of B} can be expressed as 
\begin{equation}
    \begin{aligned}
    \bm{B}^\ast = \frac{\mu(\bm{\tilde{Y} - \bm{E}_l + \bm{KJ}) + \bm{M}_2 - \bm{M}_4}}{2\mu}.
    \end{aligned}
\end{equation}

To restrict $\bm{B}^\ast$ to the feasible region, all its elements can be further projected to $[0, 1]$ as 
\begin{equation}
\setlength{\abovedisplayskip}{0pt}
\setlength{\belowdisplayskip}{0pt}
    \begin{aligned}
        \bm{B}_{ij} = \Pi(\bm{B}^\ast_{ij}),
    \end{aligned}
\label{eq: the solution to subproblem B}
\end{equation}
where the projection function $\Pi(x) = 1$ and $=0$ for $x>1$ and $x<0$, respectively, and $\Pi(x) = x$ for any $x \in[0,1]$.

\noindent
\textbf{Update $\bm{J}$:} 
The subproblem regarding $\bm{J}$ is
\begin{equation} \label{eq: the subproblem of J}
	\begin{aligned}
	 	&     \min\limits_{\bm{J}}
	            \mathrm{tr}\left(\bm{M}_3^\top\left(\bm{Z} - \bm{J}\right)\right)
	        + \mathrm{tr}\left(\bm{M}_4^\top(\bm{B} - \bm{KJ})\right)\\
		  &\ \ \ \ \ + \frac{\mu}{2}\left(\Vert\bm{Z} - \bm{J}\Vert_F^2 + \Vert\bm{B} 
		    - \bm{KJ}\Vert_F^2\right).
	\end{aligned}
\end{equation}

By computing the derivation of Eq.~\eqref{eq: the subproblem of J} {\em{w.r.t.}} $\bm{J}$ and then setting it as zero, the closed-form solution is 
\begin{equation}\label{eq: the solution to subproblem J}
\setlength{\abovedisplayskip}{0pt}
\setlength{\belowdisplayskip}{0pt}
	\begin{aligned}
		\bm{J} = \left(\bm{I} + \bm{K}^\top\bm{K} \right)^{-1}
		\left(\bm{Z} + \bm{K}^\top\bm{B} + \frac{\bm{M}_3 + \bm{K}^\top\bm{M}_4}{\mu}\right).
	\end{aligned}
\end{equation}

\noindent
\textbf{Update $\bm{K}$:} 
By dropping the unrelated terms to $\bm{K}$, we have
\begin{equation} \label{eq: the subproblem of K}
    \begin{aligned}
		&\min\limits_{\bm{K}}
		\mathrm{tr}(\bm{M}_4^\top(\bm{B} - \bm{KJ}))
		+\mathrm{tr}(\bm{M}_5^\top(\bm{X} - \bm{K})) \\
		& \ \ \ \ \ +\frac{\mu}{2}(\Vert\bm{B} - \bm{KJ}\Vert_F^2 + \Vert\bm{X} - \bm{K}\Vert_F^2).
    \end{aligned}
\end{equation}

Similarly, we compute the derivation of Eq.~\eqref{eq: the subproblem of K} {\em{w.r.t.}} $\bm{K}$ and then set it as zero, so we have
\begin{equation}\label{eq: the solution to subproblem K}
\setlength{\abovedisplayskip}{0pt}
\setlength{\belowdisplayskip}{0pt}
    \begin{aligned}
    \bm{K} = 
    \left(\frac{\bm{M}_4\bm{J}^\top + \bm{M}_5}{\mu} + \bm{B}\bm{J}^\top + \bm{X}\right)
     \left(\bm{J}\bm{J}^\top + \bm{I}\right)^{-1}.
    \end{aligned}
\end{equation}

\noindent
\textbf{Update $\bm{E}_f$ constrained by $\ell_1$-norm:}
\begin{equation}
\setlength{\abovedisplayskip}{0pt}
\setlength{\belowdisplayskip}{0pt}
	\begin{aligned}
\label{eq: update E_f_1 simply}
   \min\limits_{\bm{E}_f}
	 	\omega\Vert \bm{E}_f\Vert_1 + 
	 	\frac{1}{2}\Vert\bm{E}_f - \hat{\bm{E}_f}\Vert_F^2,
	\end{aligned}
\end{equation} 
where $\omega = \frac{\lambda_2}{\mu}$ and $\hat{\bm{E}_f} = \bm{\tilde{X}} - \bm{X} + \frac{\bm{M}_1}{\mu}$. We can extend the soft-thresholding~(shrinkage) operator introduced in~\cite{lin2010augmented,shi2014audio} to the matrix by applying it in an element-wise way, so all its elements can be updated as 
\begin{equation}\label{eq: the solution to subproblem E1}
	{[\bm{E}_f]}_{ij} = \left\{
	\begin{aligned}
		&\hat{[\bm{E}_f]}_{ij} - \omega,\ \ \ \ if\ \ \hat{[\bm{E}_f]}_{ij} > \omega,\\
		&\hat{[\bm{E}_f]}_{ij} + \omega, \ \ \ \ if\ \ \hat{[\bm{E}_f]}_{ij} < -\omega, \\
		&{0},\ \ \ \ \ \ \ \ \ \ \ \ \ otherwise,
	\end{aligned}
	\right.
\end{equation}
where $[\cdot]_{ij}$ represents the $(i,j)$-th element of the corresponding matrix.

\noindent
\textbf{Update $\bm{E}_l$:}
\begin{equation}
\setlength{\abovedisplayskip}{0pt}
\setlength{\belowdisplayskip}{0pt}
	\begin{aligned}
   \min\limits_{\bm{E}_l}
		 \xi\Vert\bm{E}_l\Vert_{2,1} +
		 \frac{1}{2} \Vert \bm{E}_l - \hat{\bm{E}_l}\Vert_F^2,
	\end{aligned}
\end{equation}
in which $\xi = \frac{\lambda_3}{\mu}$ and $\hat{\bm{E}_l} = \bm{\tilde{Y}} - \bm{B} + \frac{\bm{M}_2}{\mu}$.

As provided in~\cite{wei2019harnessing}, the closed-form solution to the general optimization problem related to $\ell_{2,1}$-norm is:

\begin{equation}\label{eq: the solution to subproblem E2}
	{[\bm{E}_l]}_i = \left\{
	\begin{aligned}
		&\frac{\Vert [\hat{\bm{E}_l}]_{i} \Vert_2-\xi}{\Vert[\hat{\bm{E}_l}]_i\Vert_2}[\hat{\bm{E}_l}]_{i},\ if\ \  \Vert [\hat{\bm{E}_l}]_i\Vert_2 > \xi,\\
		&\ \ \ \ \ \ \ \ \ 0, \ \ \ \ \ \ \ \ \ \ \ \ \ \ \ otherwise,
	\end{aligned}
	\right.
\end{equation}
where $[\cdot]_i$ represents the $i$-th row of the related matrix.

The entire optimization process for our model is summarized in Algorithm~\ref{algorithm: the algorithm for feature and label noise}.

\begin{algorithm2e}[t]
\caption{The algorithm for solving the proposed FLR}
\label{algorithm: the algorithm for feature and label noise}
\KwIn {noisy feature matrix $\bm{\tilde{X}}$, noisy label matrix $\bm{\tilde{Y}}$, \\ \ \ \ \ \ \ \ \ \ \ \ \ trade-off parameters: $\lambda_1>0$, $\lambda_2>0$, $\lambda_3>0$.}
\KwOut {optimized $\bm{X}^*$ and $\bm{Z}^*$.}
Let $\bm{X}$, $\bm{Z}$, $\bm{J}$, $\bm{E}_f$, $\bm{E}_l$, $\bm{B}$, $\bm{K}$, $\bm{M}_1$, $\bm{M}_2$, $\bm{M}_3$, $\bm{M}_4$, and $\bm{M}_5$ as zero matrices, $\mu = 10^{-3}$, $\rho = 1.2$, $\epsilon = 10^{-6}$, $iter\_max = 1000$, $iter = 0$\;
\While{not converge}{
Update $\bm{X}$ via Eq.~\eqref{eq: the solution to subproblem X}, 
$\bm{Z}$ via Eq.~\eqref{eq: the solution to subproblem Z}, 
$\bm{B}$ via Eq.~\eqref{eq: the solution to subproblem B}, 
$\bm{J}$ via Eq.~\eqref{eq: the solution to subproblem J}, 
$\bm{K}$ via Eq.~\eqref{eq: the solution to subproblem K},
$\bm{E}_f$ via Eq.~\eqref{eq: the solution to subproblem E1},
$\bm{E}_l$ via Eq.~\eqref{eq: the solution to subproblem E2}, respectively.\\
Update the multipliers\\
$\bm{M}_1 := \bm{M}_1 + \mu(\bm{\tilde{X}} - \bm{X} - \bm{E}_f)$,\\
$\bm{M}_2 := \bm{M}_2 + \mu(\bm{\tilde{Y}} - \bm{B} - \bm{E}_l)$,\\
$\bm{M}_3 := \bm{M}_3 + \mu\left(\bm{Z}-\bm{J}\right)$,\\
$\bm{M}_4 := \bm{M}_4 + \mu\left(\bm{B} - \bm{KJ}\right)$,\\
$\bm{M}_5 := \bm{M}_5 + \mu\left(\bm{X} - \bm{K}\right)$.\\
Update the parameter $\mu$ by $\mu := \rho\mu$.\\
$iter := iter + 1$.\\
Check the convergence conditions:\\
$\left\| \bm{\tilde{X}} - \bm{X} - \bm{E}_f\right\|_F \leq \epsilon $
and
$\left\|\bm{\tilde{Y}} - \bm{B} - \bm{E}_l\right\|_F \leq \epsilon$
and
$\left\|\bm{Z}-\bm{J}\right\|_F \leq \epsilon$
and
$\left\|\bm{B} - \bm{KJ}\right\|_F \leq \epsilon$
and $\left\|\bm{X} - \bm{K}\right\|_F \leq \epsilon$;
or $iter > iter\_max$.		

}
\end{algorithm2e}

\subsubsection{Computational Complexity}
Eq.~\eqref{eq: the solution to subproblem X} and Eq.~\eqref{eq: the solution to subproblem Z} in Algorithm~\ref{algorithm: the algorithm for feature and label noise} are both accomplished by SVD, and their complexities are $\mathcal{O}(\min(n^2, nd^2))$ and $\mathcal{O}(\min(n^2d, nd^2))$, respectively. Furthermore, two $d \times d$ matrices are inverted in Eq.~\eqref{eq: the solution to subproblem J} and Eq.~\eqref{eq: the solution to subproblem K}, and the complexity of these two steps is $\mathcal{O}(2d^3)$. In Eq.~\eqref{eq: the solution to subproblem E1}, one should compute the $\ell_1$-norm of a $n\times d$ matrix $\bm{E}_f$, so the complexity is $\mathcal{O}(nd)$. In Eq.~\eqref{eq: the solution to subproblem E2}, one should compute the $\ell_{2,1}$-norm of each row of a $n \times c$ matrix $\bm{E}_l$, so the complexity is $\mathcal{O}(nc)$. Thereby, the total computational complexity of Algorithm~\ref{algorithm: the algorithm for feature and label noise} is $\mathcal{O}\left(\left(\min(n^2, nd^2) + \min(n^2d, nd^2) + 2d^3 + nd +  nc\right)k \right)$ by assuming that all equations in Line 3 are iterated $k$ times. Note that the complexity of Algorithm~\ref{algorithm: the algorithm for feature and label noise} is squared to the number of training examples $n$ at most~(as lots of other robust learning approaches), so the complexity is reasonably acceptable in practical uses.

\section{Theoretical Analyses}
This section provides the theoretical analyses on FLR. Firstly, we prove that the optimization process in Algorithm~\ref{algorithm: the algorithm for feature and label noise} will converge to a stationary point and then theoretically reveal that the generalization risk of FLR is upper bounded.

\subsection{Proof of Convergence}
Note that our unified optimization objective is non-convex with multi-variables. It is difficult to guarantee the convergence of the standard ADMM optimal framework for solving multi-block problems, especially with non-convex constraints, so it is unrealistic to analyze the global optimality of our iteration algorithm. Therefore, here we investigate the local convergence of Algorithm~\ref{algorithm: the algorithm for feature and label noise}, which reveals that the iteration points converge to a stationary point under some mild conditions.

\newtheorem{theorem}{Theorem}
\newtheorem{lemma}[theorem]{Lemma}
\label{theorem: the convergence theorem of feature and label noise}
\begin{theorem}
    Let $\{\Gamma_t =$ $(\bm{X}_t, \bm{Z}_t,$ $\bm{B}_t, \bm{J}_t,$ $\bm{K}_t,$ $\bm{E}_{l, t},$ $\bm{E}_{f, t},$ $\{\bm{M}_{i,t}\}_{i=1}^5 )\}_{t = 1}^\infty$
    be the sequence generated by Algorithm~\ref{algorithm: the algorithm for feature and label noise}. 
    Assume that 
        $\lim\limits_{t \rightarrow +\infty} 
        \mu_t(\bm{K}_{t + 1} - \bm{K}_t) = 0$, 
        $\lim\limits_{t \rightarrow +\infty} 
        \mu_t(\bm{J}_{t + 1} - \bm{J}_t) = 0$, 
        $\lim\limits_{t \rightarrow +\infty}
        \mu_t(\bm{E}_{f, t + 1} - \bm{E}_{f, t}) = 0$,
        $\lim\limits_{t \rightarrow +\infty}
        \mu_t(\bm{E}_{l, t + 1} - \bm{E}_{l, t}) = 0$, and then we have
    \begin{itemize}
        \item[1.] The sequence $\{\Gamma_t\}_{t = 1}^{\infty}$ is bounded;
        \item[2.] The sequence $\{\Gamma_t\}_{t = 1}^{\infty}$ has at least one accumulation point. For any accumulation point
        $\Gamma^*$ $= (\bm{X}^*,$ $\bm{Z}^*, \bm{B}^*, \bm{J}^*, \bm{K}^*, \bm{E}_f^*, \bm{E}_l^*, \{\bm{M}_i^*\}_{i=1}^5)$,
        $(\bm{X}^*,$ $\bm{Z}^*,$ $\bm{B}^*,$ $\bm{J}^*,$ $\bm{K}^*,$ $\bm{E}_f^*,$ $\bm{E}_l^*)$ is a stationary point of the optimization Eq.~\eqref{eq: equivalent form for feature and label correction with auxiliary variables}.

    \end{itemize}
\end{theorem}

Note that the above limiting equations are actually mild conditions and are also used in~\cite{zhang2018lrr,guo2021rank}. The detailed proof is included in Appendix. Moreover, we empirically verify the convergence of our proposed algorithm by experimental results.

\subsection{Generalization Bound}
This subsection analyses the property of generalizability of our proposed FLR. 

\subsubsection{Preliminaries}
Note that the goal of FLR is to find a suitable project matrix $\bm{Z}$, given the corrupted example features $\bm{\tilde{X}}$ and noisy labels $\bm{\tilde{Y}}$.

Eq.~\eqref{eq: our proposed model for feature and label correction hard} can be rewritten to the following expression with several hard constraints, and that is 

\begin{equation}
    \begin{aligned}
        &\min_{\substack{\bm{Z},\bm{X}, \\ \bm{E}_f, \bm{E}_l}} \sum_{(i,j)} \ell\left(\left(\bm{XZ} + \bm{E}\right)_{i, j}, \bm{\tilde{Y}}_{i,j}\right)\\
	&\text{s.t.}\ \Vert\bm{X}\Vert_* \leq \mathcal{X}_*, \Vert \bm{Z} \Vert_* \leq \mathcal{Z}_*, \Vert\bm{E}_l\Vert_{2,1} \leq \mathcal{E}_{l, 21}, \\
    &\ \ \ \ \ \ \Vert\bm{\tilde{X} - \bm{X}}\Vert_1 \leq \mathcal{\bm{E}}_{f,1}, \bm{XZ}\in\left\{0,1\right\}^{n\times c},
    \end{aligned}
\end{equation}
where $\bm{E}_f = \bm{\tilde{X} - \bm{X}}$ and $(i, j)\in \{1, \dots, n\}\times\{1, \dots, c\}$. 
The decision function $f_\theta: \mathcal{\tilde{X}} \rightarrow \mathcal{\tilde{Y}}$, and $f_\theta(i,j)$ $= (\bm{\tilde{X}}_i - \bm{E}_{fi})\bm{Z}\bm{I}^j + \bm{E}_{ij}$, which is controlled by the feasible solution $\theta$ $= (\bm{X},$ $\bm{Z},$ $\bm{E}_f,$ $\bm{E}_l)$. The feasible solution set $\Theta \!=\! \left\{(\bm{X},\! \bm{Z},\! \bm{E}_f,\! \bm{E}_l)| \Vert\bm{X}\Vert_* \!\leq\! \mathcal{X}_*, \Vert \bm{Z} \Vert_* \!\leq\! \mathcal{Z}_*, \Vert\bm{E}_l\Vert_{2,1}\! \leq\! \mathcal{E}_{l, 21}, \!\Vert\bm{\tilde{X}\! -\! \bm{X}}\Vert_F \leq \mathcal{\bm{E}}_{f,F}, \bm{XZ}\in\left\{0,1\right\}^{n\times c}\right\}$ and the set of feasible functions $\mathcal{F}_\Theta = \{f_\theta | \theta \in \Theta\}$. $\bm{I}^j$ is the $j$-th column of identity matrix  $\bm{I}\in \mathbb{R}^{c \times c}$, and $\bm{X}_i$ is the $i$-th row of matrix $\bm{X}$ in general.

Furthermore, we introduce the following two ``$\ell$-risk'' quantities:
\begin{itemize}
    \item Expected ``$\ell$-risk'': $R_\ell(f) = \mathbb{E}_{i,j} \left[\ell\left(f(i,j), \bm{\tilde{Y}}_{i,j}\right)\right]$;
    \item Empirical ``$\ell$-risk'': $\hat{R}_\ell(f) = \frac{1}{n_r}\sum_{(i,j)} \ell\left(f(i,j), \bm{\tilde{Y}}_{i,j}\right)$,
\end{itemize}
where $n_r$ is the number of observed entries. Therefore, the proposed FLR is designed to obtain a proper $\theta^*$ that parameterizes the optimal $f_\theta^* = \arg\min_{f \in \mathcal{F}_\Theta}\hat{R}_\ell(f)$.

\subsubsection{Generalization Bound of FLR}

The Rademacher complexity quantitatively measures the diversity of a function class, which is a useful tool for analyzing the bound of the generalization error of the learning algorithm. Specifically, we link the quality of training features and labels to Rademacher complexity, and show that the high-quality features and labels will result in the lower model complexity and thus a smaller error bound. To this end, we denote that $\mathcal{R}_n(\mathcal{F}) := \mathbb{E}[\mathcal{R(\mathcal{F})}]\nonumber$ as the Rademacher complexity of the function class $\mathcal{F}$, and $\mathcal{R(\mathcal{F})} := \mathbb{E}_\sigma[\sup\limits_{f\in\mathcal{F}}\frac{1}{n_r}\sum\limits_{\alpha = 1}^{n_r}\sigma_\alpha f(\alpha)]\nonumber$ as the empirical Rademacher complexity on the training set, where $\sigma_i \in \{-1, 1\}$~($i = 1, 2, \dots, n_r$) are $i.i.d.$ Rademacher random variables.

\begin{theorem}
\label{theorem: the upper bound of rademacher complexitys}
    Let $\ell$ be the loss function bounded by $\mathcal{B}$ with Lipschitz constant $L_\ell$, and $\delta$ be a constant where $0 \leq \delta \leq 1 $. Then with probability at least $1 - \delta$, we have 
    \begin{equation}
    \setlength{\abovedisplayskip}{0pt}
\setlength{\belowdisplayskip}{0pt}
        \begin{aligned}
            \max\limits_{f \in \mathcal{F}_\Theta}|R_l(f) - \hat{R}_l(f)| \leq 2 L_\ell \mathcal{R}_n(\mathcal{F}_\Theta) + \mathcal{B}\sqrt{\frac{\ln{1/\delta}}{2nc}},
        \end{aligned}
    \end{equation}
    in which
\begin{equation}
    \begin{aligned}
    \mathcal{R}_n(\mathcal{F}_\Theta)
    & \leq 
    \mathcal{E}_{l,21}\mathcal{C}_1 + \min\left\{\mathcal{X}_*\mathcal{Z}_*\mathcal{C}_2, \mathcal{Z}_*(\mathcal{\tilde{X}}_F + \sqrt{d}\mathcal{E}_{f,1})\mathcal{C}_3, \mathcal{C}_4\right\},
    \end{aligned}
    \label{eq: the upper bound of rademacher complexity for feature and label noise}
\end{equation}
with $\mathcal{C}_1= \sqrt{\frac{3\ln{c}}{nc}}$, $\mathcal{C}_2 = \sqrt{\frac{\ln{(2n_c)}}{nc}}$, $\mathcal{C}_3 = \frac{1}{\sqrt{nc}}$ and $\mathcal{C}_4 = \sqrt{\frac{2}{c}}$.
\end{theorem}

The proof of Theorem~\ref{theorem: the upper bound of rademacher complexitys} is provided in Appendix. As mentioned before, the matrix $\bm{E}_f$ captures the feature noise, and its $\ell_{1}$-norm value is upper bounded by $\mathcal{E}_{f,1}$. Also, the label noise matrix $\bm{E}_l$ model is upper bounded by $\mathcal{E}_{l,21}$. The recovered clean feature matrix $\bm{X}$ is assumed to be sparse. Specifically, the $\mathcal{E}_{f,1}$ and $\mathcal{E}_{l,21}$ are governed by the severity of feature noise and label noise, respectively. The light noise will lead to small $\mathcal{E}_{f,1}$ and $\mathcal{E}_{l,21}$, which can further reduce the upper bound of the expected $\ell$-risk in the right-hand side of Eq.~\eqref{eq: the upper bound of rademacher complexity for feature and label noise}. Conclusively, the high-quality features and labels of training examples will contribute to a lower model complexity and thus ensure a smaller error bound.

\begin{table}[htb]
\setlength{\belowcaptionskip}{3pt}
\caption{The comparison of mean test accuracy~(\%) of various methods on four UCI benchmark datasets. The highest and second highest records are \textbf{bolded} and \underline{underlined}, respectively.}
\resizebox{1\linewidth}{!}{
\begin{tabular}{l|lll|lll|lll|lll}
\hline
Gaussian Noise      & \multicolumn{3}{c|}{Glass ($n = 214, d = 9, c= 6$)}                                 & \multicolumn{3}{c|}{Wine ($n = 178, d = 13, c= 3$)}                                 & \multicolumn{3}{c|}{CNAE9 ($n = 1080, d = 856, c= 9$)}                                    & \multicolumn{3}{c}{Waveform($n = 5000, d = 21, c= 3$)}                                 \\ \hline
($\sigma_f$, $\eta_l$)  & (0.2, 0.3)         & (0.2, 0.6)        & (0.5, 0.6)        & (0.2, 0.3)        & (0.2, 0.6)        & (0.5, 0.6)        & (0.2, 0.3)         & (0.2, 0.6)         & (0.5, 0.6)          & (0.2, 0.3)         & (0.2, 0.6)         & (0.5, 0.6)         \\ \hline
RPCA~\cite{candes2011robust}                
& 45.25$\pm$3.12  & 35.45$\pm$1.32 & 31.47$\pm$2.43 
& 85.65$\pm$2.11  & 73.41$\pm$1.78 & 67.41$\pm$2.13 
& 65.25$\pm$1.21  & 50.12$\pm$1.76 & 28.45$\pm$1.33   
& 55.15$\pm$1.12 & 50.12$\pm$1.76 & 40.40$\pm$1.24 \\
GCE~\cite{zhang2018generalized}                 
& 24.55$\pm$13.09  & 26.37$\pm$17.72  & 23.64$\pm$13.78 
& 40.00$\pm$17.30  & 40.00$\pm$16.85  & 36.67$\pm$21.37 
& 49.26$\pm$25.75  & 29.07$\pm$13.30  & 18.52$\pm$4.49   
& 62.00$\pm$9.73  & 56.04$\pm$8.67   &  55.08$\pm$8.45  \\
Co-teaching~\cite{han2018co}         
& \underline{57.67$\pm$4.26}  & 38.53$\pm$9.18   & \underline{48.22$\pm$6.12} 
& 89.99$\pm$5.05  & 57.78$\pm$14.62  & 48.33$\pm$21.57 
& 76.60$\pm$2.80  & 50.19$\pm$2.39   & 30.62$\pm$6.93 
& 81.59$\pm$0.55  & \underline{71.55$\pm$1.31}   & \underline{70.26$\pm$2.76}  \\
JoCoR~\cite{wei2020combating}               
& 37.36$\pm$2.45  & 26.36$\pm$4.21  & 27.67$\pm$5.57 
& 83.33$\pm$3.93  & 44.44$\pm$3.40  & 43.89$\pm$10.28 
& 43.30$\pm$0.70  & 30.02$\pm$2.15  & 21.40$\pm$2.77   
& 56.57$\pm$2.00  & 40.19$\pm$1.33  & 39.43$\pm$3.35  \\
$\mathcal{L}_{DMI}$~\cite{xu2019l_dmi} 
& 54.55$\pm$3.21  & \underline{39.09$\pm$6.10}  & 24.55$\pm$13.09 
& \underline{90.00$\pm$8.24}  & \underline{76.67$\pm$17.74} & \underline{76.67$\pm$7.24} 
& \underline{88.33$\pm$2.82}  & \underline{69.63$\pm$8.37}  & \underline{60.00$\pm$4.87}   
& \textbf{84.96$\pm$2.23}  & 69.68$\pm$3.99  & 57.36$\pm$14.21  \\
LQF~\cite{zhu2022beyond}                 
& 43.64$\pm$10.47  &37.27$\pm$6.74  & 34.54$\pm$5.18 
& 89.82$\pm$7.86   &74.44$\pm$19.48 & 68.89$\pm$12.79 
& 49.63$\pm$14.07   & 30.92$\pm$10.85  & 13.15$\pm$2.65   
& 80.36$\pm$3.69  & 67.92$\pm$1.61  & 61.48$\pm$2.15  \\
\textbf{FLR} \footnotesize{(ours)}       
& \textbf{68.37$\pm$4.82}  & \textbf{66.98$\pm$6.45} & \textbf{51.43$\pm$2.13} 
& \textbf{96.57$\pm$1.27} & \textbf{84.57$\pm$6.26} & \textbf{78.28$\pm$1.57} 
& \textbf{89.72$\pm$1.10}  & \textbf{76.57$\pm$3.58}  & \textbf{67.49$\pm$1.60}   
& \underline{84.48$\pm$1.05}  & \textbf{76.08$\pm$1.13}  & \textbf{71.98$\pm$1.06}  \\ \hline
Laplacian Noise     &                    &                   &                   &                   &                   &                   &                    &                    &                     &                    &                    &                    \\ \hline
RPCA~\cite{candes2011robust}                
& 45.25$\pm$3.12  & \underline{39.21$\pm$2.56}  & 32.74$\pm$3.65 
& 92.25$\pm$0.12  & 75.20$\pm$2.50  & 62.04$\pm$3.55 
& 62.52$\pm$3.11  & 45.52$\pm$2.05  & 29.24$\pm$1.10  
& 52.25$\pm$0.11  & 55.25$\pm$5.02  & 47.42$\pm$2.30  \\
GCE~\cite{zhang2018generalized}                 
&  23.64$\pm$12.61   &  25.45$\pm$12.28  &  21.82$\pm$14.15
&  40.00$\pm$22.01   &  28.89$\pm$19.40    & 31.11$\pm$21.37 
&  47.59$\pm$24.04   &  24.44$\pm$9.32   &  17.41$\pm$6.33   
&  59.68$\pm$7.06    &  55.84$\pm$8.39    &  51.28$\pm$7.71  \\
Co-teaching~\cite{han2018co}         
& \underline{58.60$\pm$4.50}  & 38.53$\pm$3.49  & \underline{35.63$\pm$2.99} 
& 91.67$\pm$1.96  & 45.00$\pm$18.15  & 44.45$\pm$5.20 
& 71.98$\pm$0.86  & 44.54$\pm$6.12  & 20.06$\pm$5.79   
& 80.96$\pm$0.48  & \underline{72.64$\pm$1.34}  & \underline{69.15$\pm$2.85}  \\
JoCoR~\cite{wei2020combating}               
& 38.68$\pm$4.16  & 28.22$\pm$4.70  & 26.12$\pm$5.75 
& 83.33$\pm$4.39 & 52.78$\pm$14.30 & 38.33$\pm$11.69 
& 38.86$\pm$1.16  & 27.95$\pm$3.44  & 16.37$\pm$1.68   
& 55.93$\pm$2.86  & 40.89$\pm$3.13  & 35.68$\pm$4.01  \\
$\mathcal{L}_{DMI}$~\cite{xu2019l_dmi} 
& 44.55$\pm$13.41  & 24.55$\pm$19.97  & 28.18$\pm$10.85 
& 92.22$\pm$8.42  & \underline{76.66$\pm$13.26}  & 61.11$\pm$13.03 
& \textbf{85.37$\pm$2.48}  & \underline{67.59$\pm$5.36}  & \underline{51.67$\pm$4.87}   
& \underline{81.72$\pm$2.33}  & 65.76$\pm$3.93  & 60.44$\pm$3.12  \\
LQF~\cite{zhu2022beyond}                 
&38.18$\pm$8.26  & 29.38$\pm$5.71  & 29.09$\pm$11.85 
&\textbf{93.33$\pm$4.65}  & 70.00$\pm$15.52  &\underline{67.78$\pm$14.38} 
& 53.15$\pm$6.27  & 26.85$\pm$9.49  & 12.22$\pm$2.49   
& 78.48$\pm$4.56  & 60.92$\pm$18.63  & 52.04$\pm$11.26  \\
\textbf{FLR} \footnotesize{(ours)}                
& \textbf{68.37$\pm$4.82}  & \textbf{50.48$\pm$1.06}  & \textbf{47.14$\pm$1.99} 
& \underline{92.56$\pm$2.56}           & \textbf{81.71$\pm$3.26}  & \textbf{74.86$\pm$4.70}   
& \underline{84.91$\pm$2.31}  & \textbf{68.60$\pm$1.31}  & \textbf{53.33$\pm$3.50}   
& \textbf{82.16$\pm$2.10}  & \textbf{74.60$\pm$0.10}  & \textbf{74.68$\pm$1.04}  \\ \hline
\end{tabular}}
\label{tab: UCI experimental results}
\end{table}

\section{Experimental Results}
In this section, we show the experimental results on various datasets to validate the effectiveness of our proposed FLR. In detail, we first compare our FLR with existing feature noise learning and label noise learning methods in different noisy setting cases. Our experimental data includes four classical UCI benchmark datasets and the popular CIFAR-10 and CIFAR-10N datasets. In our experiments, the training data is corrupted~(with both the feature noise and label noise) while the test data is clean.

\subsection{Experiments on UCI Benchmark Datasets}
We compare FLR with six baseline algorithms on four UCI benchmark datasets, including Glass, Wine, CNAE9, and Waveform. Note that 80\% of the examples in each dataset are randomly chosen to establish the training set, and the remaining 20\% ones are severed as the test set. To incorporate different levels of hybrid noise into training sets, we add the Gaussian noise or Laplacian noise under different standard deviations $\sigma_f$ with mean $\mu = 0$ on features of examples. Meanwhile, we randomly pick up 0\%, 30\%, 60\% examples from the training sets and inject symmetric label noise~\cite{patrini2016lossFactorizationWeaklySupervisedLearning} to these selected examples, and the noise rate is marked as $\eta_l$. Such contamination and partition are conducted five times, so the accuracies are the mean values of the outputs of five independent trials.

Our compared baseline methods include RPCA + a classifier~\cite{candes2011robust}, GCE~\cite{zhang2018generalized}, Co-teaching~\cite{han2018co}, JoCoR~\cite{wei2020combating}, $\mathcal{L}_{DMI}$~\cite{xu2019l_dmi}, LQF~\cite{zhu2022beyond}, and RRL~\cite{li2021learning}. Note that the first approach is designed for feature noise learning~(it needs an additional classifier for the classification task), and the following six focus on label noise learning. Furthermore, RRL is designed for image tasks with data noise, so it is not suitable for UCI datasets.

The classification accuracies of all compared approaches on the test set are listed in Table~\ref{tab: UCI experimental results}. We observe that FLR yields better performance than other baseline methods in most cases or achieves the similar performance to them. Especially with increasing levels of hybrid noise, the accuracies of most compared methods decrease correspondingly, but FLR still performs robustly in those cases.

\begin{figure*} [t] 
	\begin{minipage}{1\linewidth}
        \captionsetup{font={scriptsize}} 
		\centering
		\subfloat[]{\includegraphics[width=0.3\linewidth]{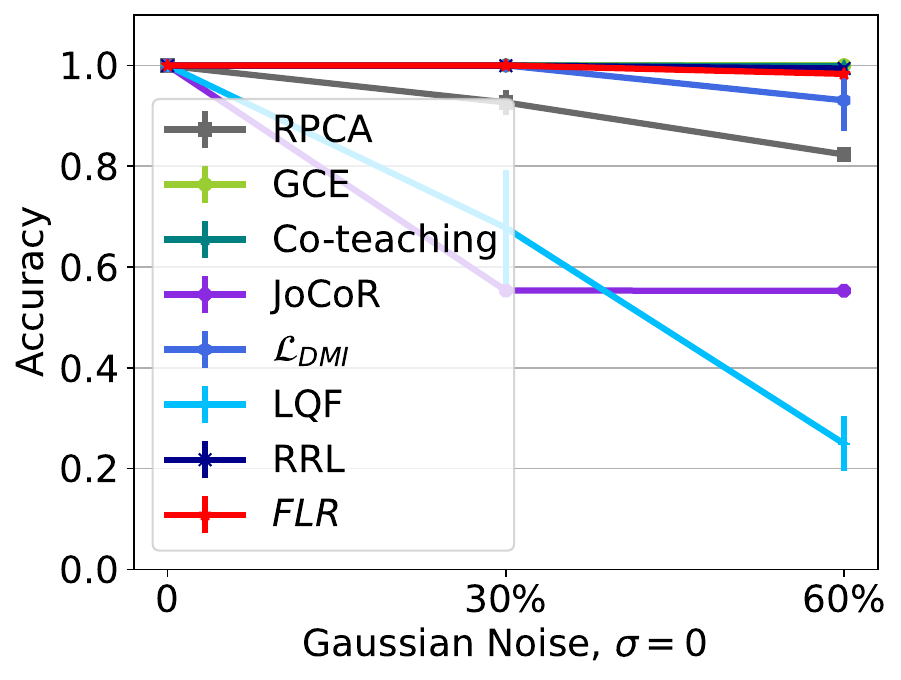}}
		\subfloat[]{\includegraphics[width=0.3\linewidth]{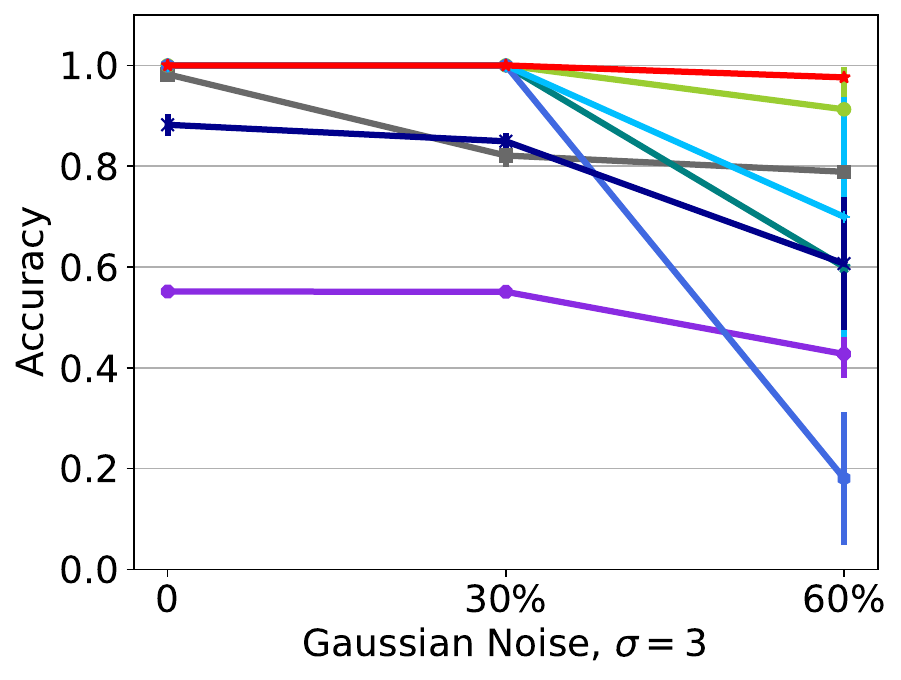}}
		\subfloat[]{\includegraphics[width=0.3\linewidth]{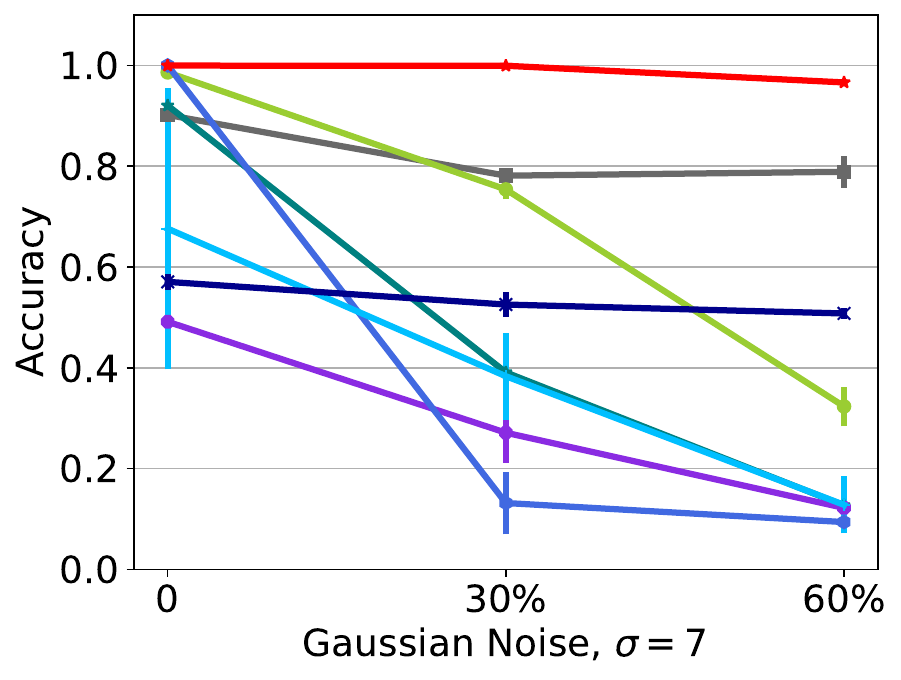}}
            \\
            \subfloat[]{\includegraphics[width=0.3\linewidth]{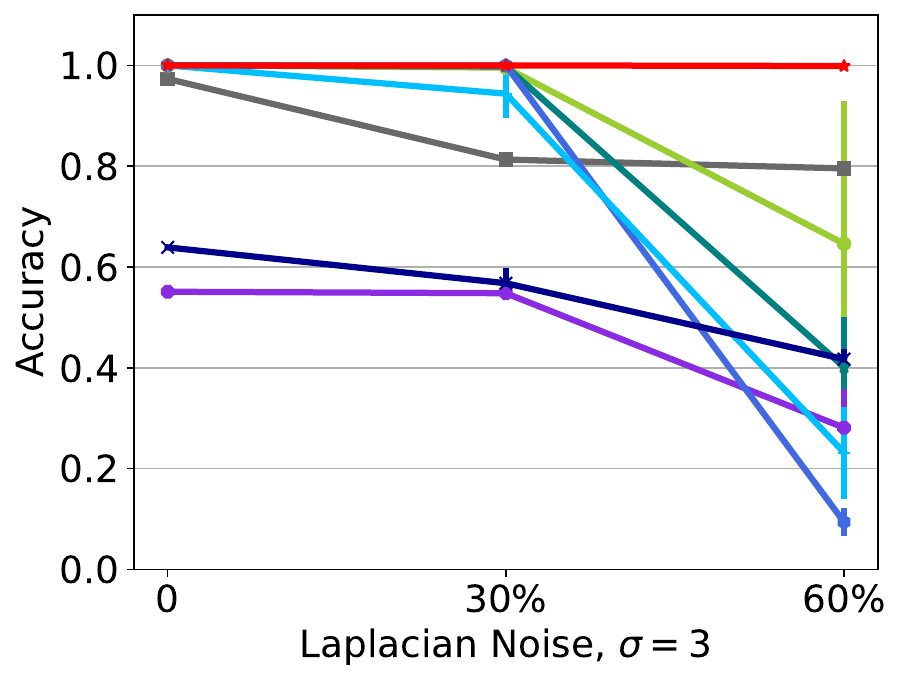}}
            \subfloat[]{\includegraphics[width=0.3\linewidth]{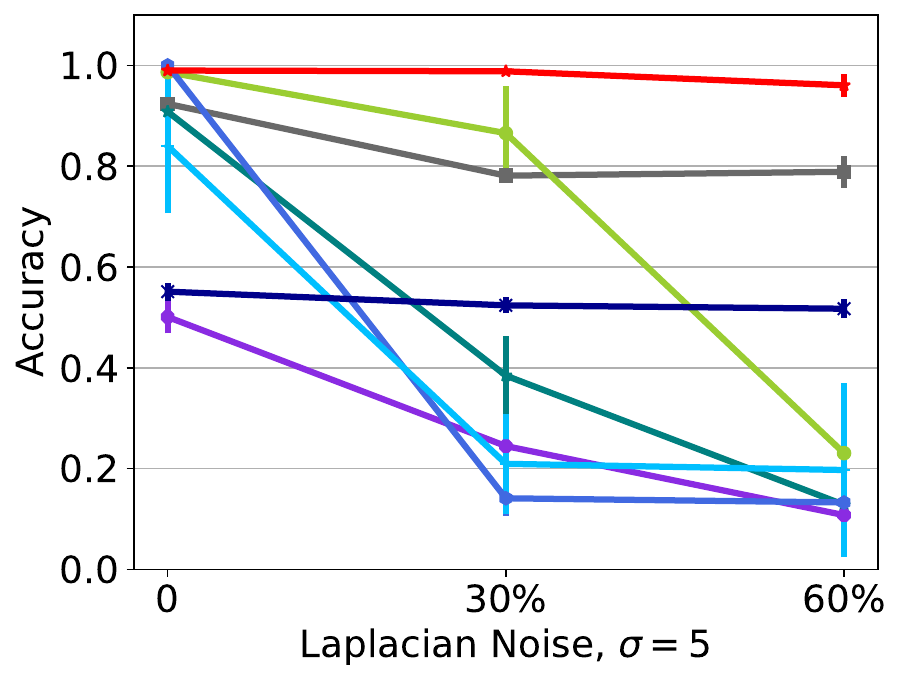}}
	\end{minipage}
	\caption{The experimental results on CIFAR-10 dataset in various noise scenarios.}
\label{fig: the cifar10 on gaussian and laplacian}
\end{figure*}

\begin{table}[t]
\centering
\caption{The comparison of mean test accuracy~(\%) of various methods on CIFAR-10N dataset with the Gaussian noise. The highest and second highest records are \textbf{bolded} and \underline{underlined}, respectively.
}
\resizebox{.8\columnwidth}{!}{
\begin{tabular}{lllll}
\hline
& Datasets & C-10N Aggre.   & C-10N Random1  & C-10N Worst \\ \hline
& RPCA~\cite{candes2011robust}              
&90.24$\pm$0.78    &82.12$\pm$2.23      &78.89$\pm$4.52             \\
& GCE~\cite{zhang2018generalized}                
&88.88$\pm$4.23    &86.78$\pm$1.89 &79.97$\pm$8.55             \\
& Co-teaching~\cite{han2018co}        
& 94.03$\pm$0.06    &94.14$\pm$0.05      &90.16$\pm$1.83\\
$\bm{\sigma_f}=3$ 
& JoCoR~\cite{wei2020combating}              
&52.94$\pm$0.42    &52.94$\pm$0.38      &49.04$\pm$0.33           \\
& $\mathcal{L}_{DMI}$~\cite{xu2019l_dmi}
&\textbf{96.03$\pm$0.08}    &95.75$\pm$0.23      &47.85$\pm$22.49            \\
& LQF~\cite{zhu2022beyond}                
&94.23$\pm$2.78    &\underline{95.82$\pm$0.23}      &\underline{92.71$\pm$4.40}           \\
& RRL~\cite{li2021learning} 
& 89.70$\pm$1.44 & 85.00$\pm$1.18 & 75.62$\pm$0.89\\
& \textbf{FLR} \footnotesize{(ours)}                
&\underline{95.70$\pm$0.15}  &\textbf{96.05$\pm$0.05}  &\textbf{94.92$\pm$0.06} \\ \hline
& RPCA~\cite{candes2011robust}             
&68.47$\pm$0.78    &69.12$\pm$1.75      &48.59$\pm$3.72          \\
& GCE~\cite{zhang2018generalized}              
&55.78$\pm$9.42    &56.40$\pm$8.52 &42.45$\pm$5.67             \\
& Co-teaching~\cite{han2018co}       
&78.68$\pm$5.32    &\underline{70.12$\pm$5.41}      &19.43$\pm$4.31\\
$\bm{\sigma_f}=7$ 
& JoCoR~\cite{wei2020combating}             
&52.91$\pm$0.50    &52.49$\pm$0.21      &42.08$\pm$0.76        \\
& $\mathcal{L}_{DMI}$~\cite{xu2019l_dmi}
&\underline{91.93$\pm$6.74}    &42.05$\pm$22.82      &34.43$\pm$11.29             \\
& LQF~\cite{zhu2022beyond}               
&82.42$\pm$6.74    &67.30$\pm$13.17 &8.97$\pm$1.97          \\
&RRL~\cite{li2021learning}
& 61.63$\pm$1.08 & 55.93$\pm$0.80 & \underline{50.98$\pm$1.12} \\
& \textbf{FLR} \footnotesize{(ours)}              
&\textbf{95.93$\pm$0.06} &\textbf{93.75$\pm$0.10} &\textbf{95.08$\pm$0.06} \\ \hline
\end{tabular}}
\label{tab: results on cifar10N agg random1 worst with gaussain}
\end{table}

\subsection{Experiments on CIFAR-10 Dataset}
To further evaluate the effectiveness of FLR, we conduct experiments on the CIFAR-10 dataset, which contains 60,000 natural images across 10 classes. In our experiments, we follow the common practice~\cite{ke2020laplacian,luo2021bi} to randomly pick up 6,000 image examples as our experimental data. The resolution of each image is $32\times32$. We extract the CNN features for each image, which are obtained from the fully connected layer of a pre-trained ResNet-18 with 512-dimensional features. The experimental settings are similar to those of UCI datasets. 

From Figure~\ref{fig: the cifar10 on gaussian and laplacian}, we can observe that FLR performs significantly better than baseline methods, especially when the training examples are heavily corrupted by hybrid noise. This is because FLR integrates the feature noise and label noise into a natural framework, so that we can successfully deal with the hybrid noise by merely using a single training process.

\begin{table}[t]
\centering
\caption{The comparison of mean test accuracy~(\%) of various methods on CIFAR-10N dataset with the Laplacian Noise. The highest and second highest records are \textbf{bolded} and \underline{underlined}, respectively.}
\resizebox{.8\columnwidth}{!}{
\begin{tabular}{lllll}
\hline
& Datasets    & C-10N Aggre. & C-10N Random1 & C-10N Worst \\ \hline
& RPCA~\cite{candes2011robust}        
&89.12$\pm$0.28   &84.01$\pm$2.75 &72.21$\pm$1.53       \\
& GCE~\cite{zhang2018generalized}                  
&85.68$\pm$1.68   &85.08$\pm$2.41 &70.13$\pm$8.61            \\
& Co-teaching~\cite{han2018co}       
&95.62$\pm$0.32   &95.33$\pm$0.17 &\underline{77.42$\pm$5.57} \\
$\sigma_f=3$ 
& JoCoR~\cite{wei2020combating}                
&52.91$\pm$0.50   &52.49$\pm$0.21 &42.08$\pm$0.76          \\
& $\mathcal{L}_{DMI}$~\cite{xu2019l_dmi}  
&92.35$\pm$6.11   &\textbf{96.10$\pm$0.09} 
&19.12$\pm$9.70          \\
& LQF~\cite{zhu2022beyond}                  
&\underline{95.66$\pm$0.45}   &95.70$\pm$0.64 &66.02$\pm$15.33           \\
& RRL~\cite{li2021learning} 
&65.40$\pm$0.85 &65.37$\pm$1.32 &59.78$\pm$1.60\\
& \textbf{FLR} (ours)                  
&\textbf{95.78$\pm$0.05}   &\underline{95.70$\pm$0.18} &\textbf{94.12$\pm$0.10} \\ \hline
& RPCA~\cite{candes2011robust}                 
&68.72$\pm$0.65   &\underline{65.22$\pm$1.15} &48.21$\pm$1.75            \\
& GCE~\cite{zhang2018generalized}                 
&59.38$\pm$7.47   &53.40$\pm$14.81 &36.92$\pm$8.91            \\
& Co-teaching~\cite{han2018co}          
&46.17$\pm$10.73   &30.75$\pm$8.15 &13.38$\pm$1.63 \\
$\sigma_f=5$ 
&JoCoR~\cite{wei2020combating}        
&25.77$\pm$7.52   &22.94$\pm$4.14 &12.81$\pm$1.53      \\
& $\mathcal{L}_{DMI}$~\cite{xu2019l_dmi} 
&\underline{72.70$\pm$24.25}   &52.40$\pm$22.15 &21.07$\pm$18.04      \\
& LQF~\cite{zhu2022beyond}                
&52.98$\pm$8.71   &51.83$\pm$16.16 &12.70$\pm$5.17             \\
& RRL~\cite{li2021learning}
&51.87$\pm$1.62 &50.37$\pm$1.11 &\underline{50.05$\pm$0.44} \\
& \textbf{FLR} (ours)             
&\textbf{76.00$\pm$0.31}   &\textbf{78.85$\pm$0.39} &\textbf{61.23$\pm$4.00}  \\ \hline
\end{tabular}}
\label{tab: results on cifar10N agg random1 worst with laplacian}
\end{table}

\subsection{Experiments on CIFAR-10N Datasets}
We compare FLR with the above baseline methods on the practical noisy dataset CIFAR-10N. Here CIFAR-10N consists of five inherently noisy label sets, among which the noise rate of ``Random1'', ``Aggre.'', and ``Worst'' are 17.23\%, 9.03\%, and 40.21\%, respectively. We use three subsets of ``Aggre.'', ``Random1'' and ``Worst'' for our experiments, which contain 10,000 examples selected from their training sets and 2,000 examples from their test sets, respectively. We also extract the CNN features for each image.

All experimental settings are similar to those above. As labels in CIFAR-10N are inherently noisy, we need to artificially introduce the feature noise to training data. The average accuracy over three independent trials is reported. The experimental results under Gaussian noise are shown in Table~\ref{tab: results on cifar10N agg random1 worst with gaussain} and experimental results under Laplacian noise are provided in Table~\ref{tab: results on cifar10N agg random1 worst with laplacian}.

From Table~\ref{tab: results on cifar10N agg random1 worst with gaussain} and Table~\ref{tab: results on cifar10N agg random1 worst with laplacian}, we can clearly observe that when the feature noise is not heavy, FLR is comparable with other baseline methods. Furthermore, FLR can achieve much more robust and satisfactory performances when the $\sigma_f$ is relatively large, while all baseline methods perform unsatisfactorily. Consequently, the existing methods cannot handle hybrid noise well, yet FLR is indeed an effective solution to tackling the hybrid noise.

\begin{figure} [h!]
	\begin{minipage}{1\linewidth}
        \captionsetup{font={scriptsize}} 
		\centering
		\subfloat[]{\includegraphics[width=0.43\linewidth]{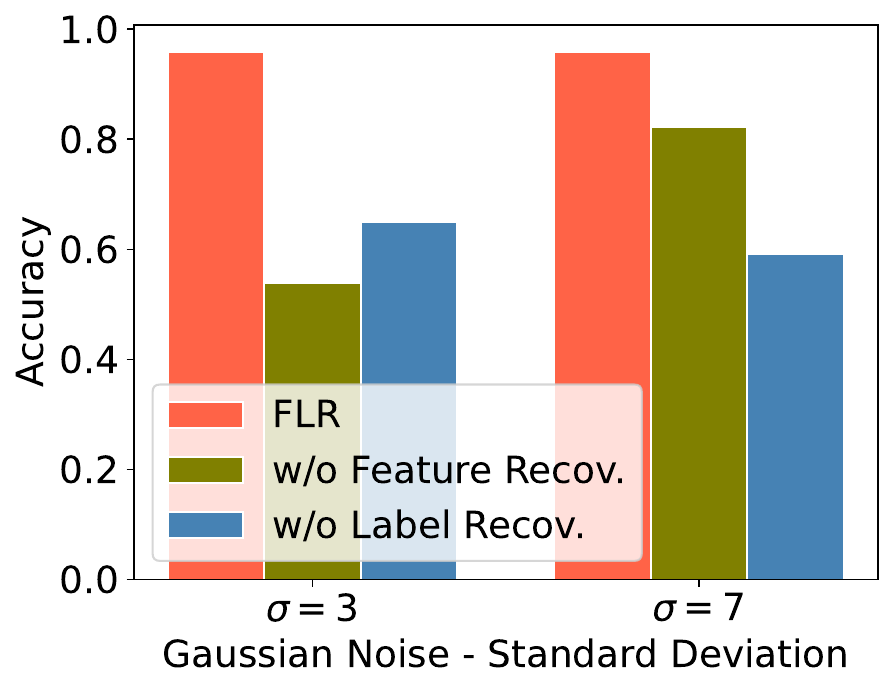}}
		\subfloat[]{\includegraphics[width=0.43\linewidth]{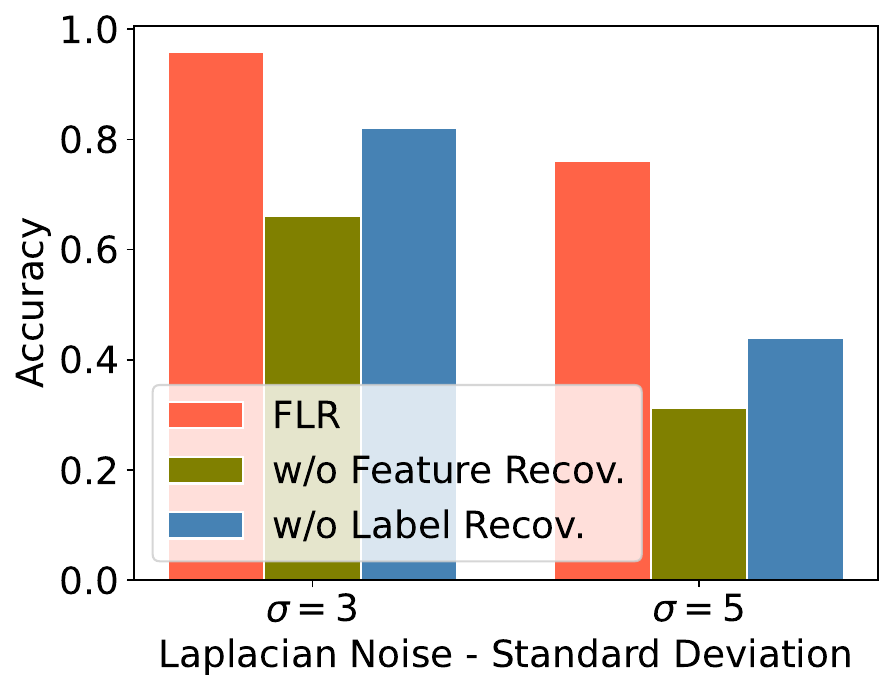}}
	\end{minipage}
	\caption{Ablation of FLR on ``Aggre.''.}
\label{fig: Ablation of FLR on ``Aggre.''.}
\end{figure}

\subsection{Ablation Study}

This section investigates the usefulness of each key component in FLR. We study the performances of three different settings on the CIFAR-10N ``Aggre.'' dataset. First, both the two terms $\bm{E}_f$ and $\bm{E}_l$ are reserved to constitute the original model~(abbreviated as ``FLR''); second, the term $\bm{E}_f$ is removed, which means that the feature recovery process is ignored~(abbreviated as ``w/o feature recov.''); third, we remove the term $\bm{E}_l$, so that the label recovery is abandoned~(abbreviated as ``w/o label recov.'').

The experimental results of these models with different feature noise are illustrated in Figure~\ref{fig: Ablation of FLR on ``Aggre.''.}. The results clearly reveal that the regular setting~(namely, FLR) achieves the best performance on the ``Aggre.'' dataset. By contrast, the accuracy will drop a lot without any of the two terms~(\emph{i.e.}, term $\bm{E}_f$ and term $\bm{E}_l$). Therefore, these two noise capture terms are essential to boost the performance of FLR.

\begin{figure} [t] 
	\begin{minipage}{1\linewidth}
        \captionsetup{font={scriptsize}} 
		\centering
		\subfloat[\begin{scriptsize}
		    $\lambda_1$, Gaussian Noise
		\end{scriptsize}]{\includegraphics[width=0.3\linewidth]{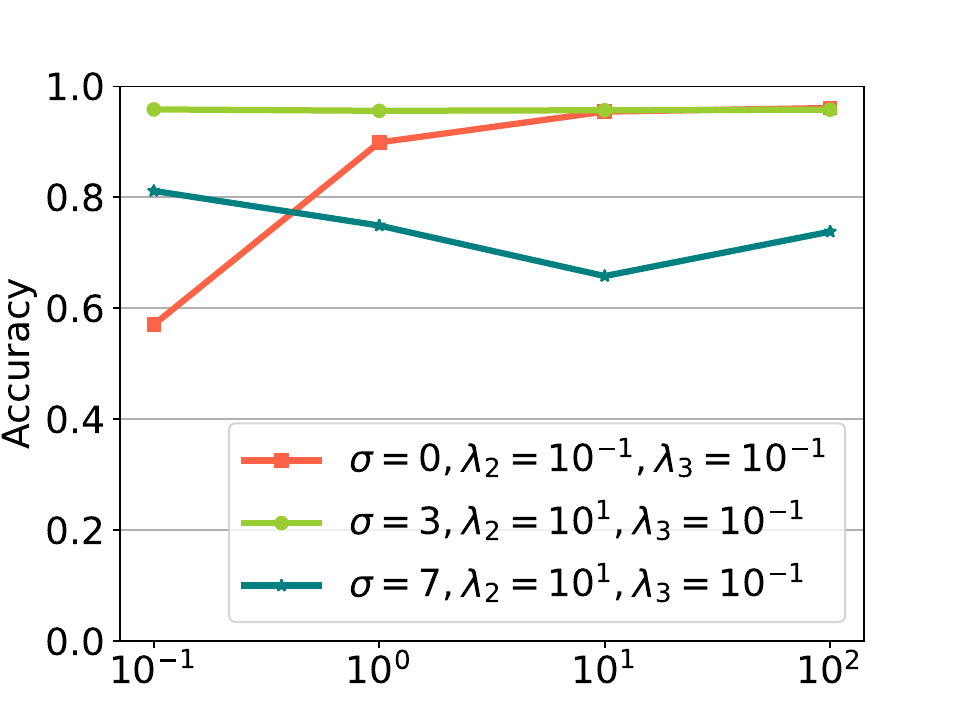}}
		\subfloat[\begin{scriptsize}
		    $\lambda_2$, Gaussian Noise
		\end{scriptsize}]{\includegraphics[width=0.3\linewidth]{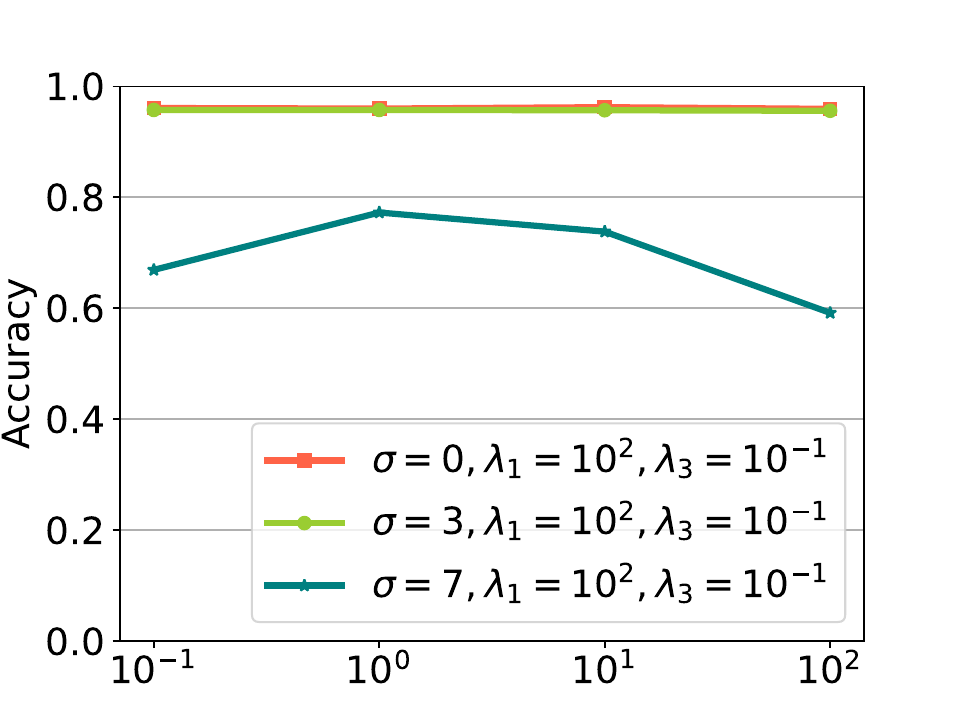}}
		\subfloat[\begin{scriptsize}
		    $\lambda_3$, Gaussian Noise
		\end{scriptsize}]{\includegraphics[width=0.3\linewidth]{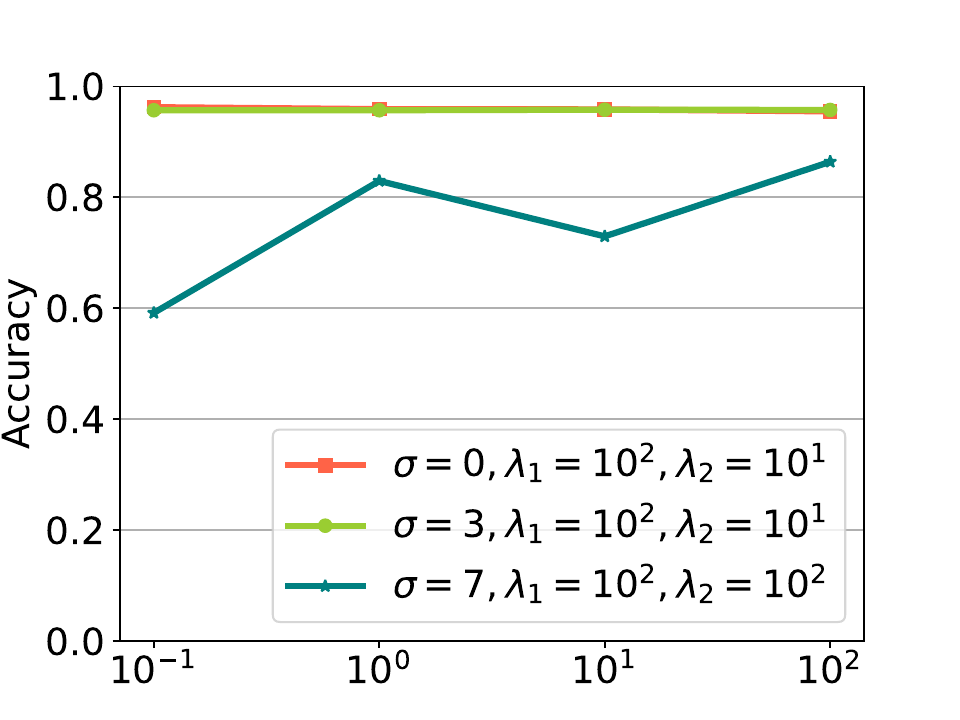}}
  	    \\
            \subfloat[\begin{scriptsize}
		    $\lambda_1$, Laplacian Noise
		\end{scriptsize}]{\includegraphics[width=0.3\linewidth]{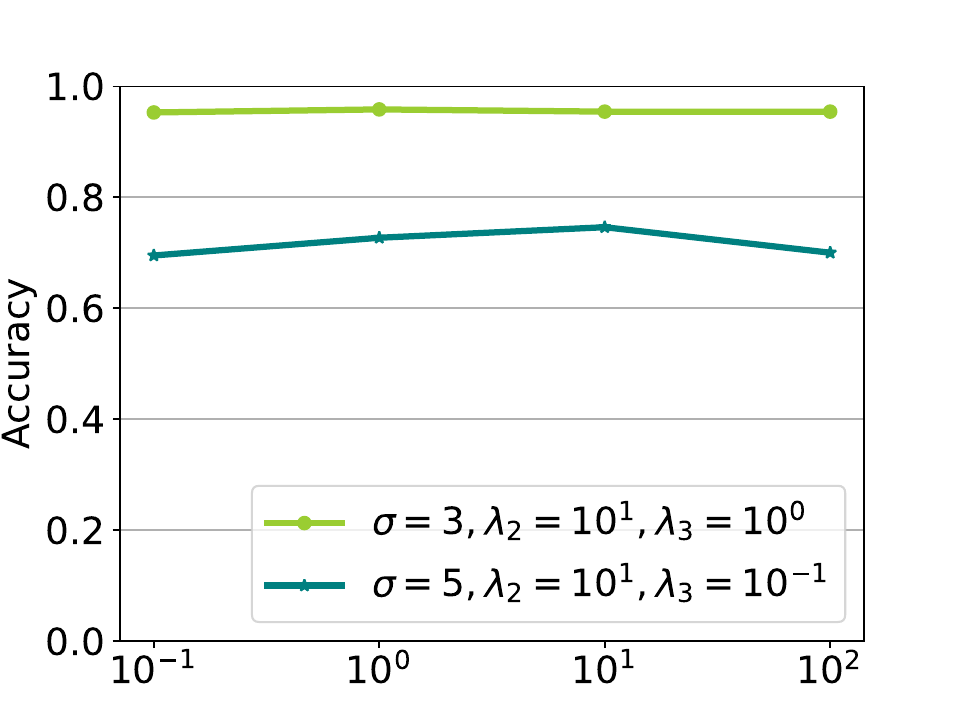}}
		\subfloat[\begin{scriptsize}
		    $\lambda_2$, Laplacian Noise
		\end{scriptsize}]{\includegraphics[width=0.3\linewidth]{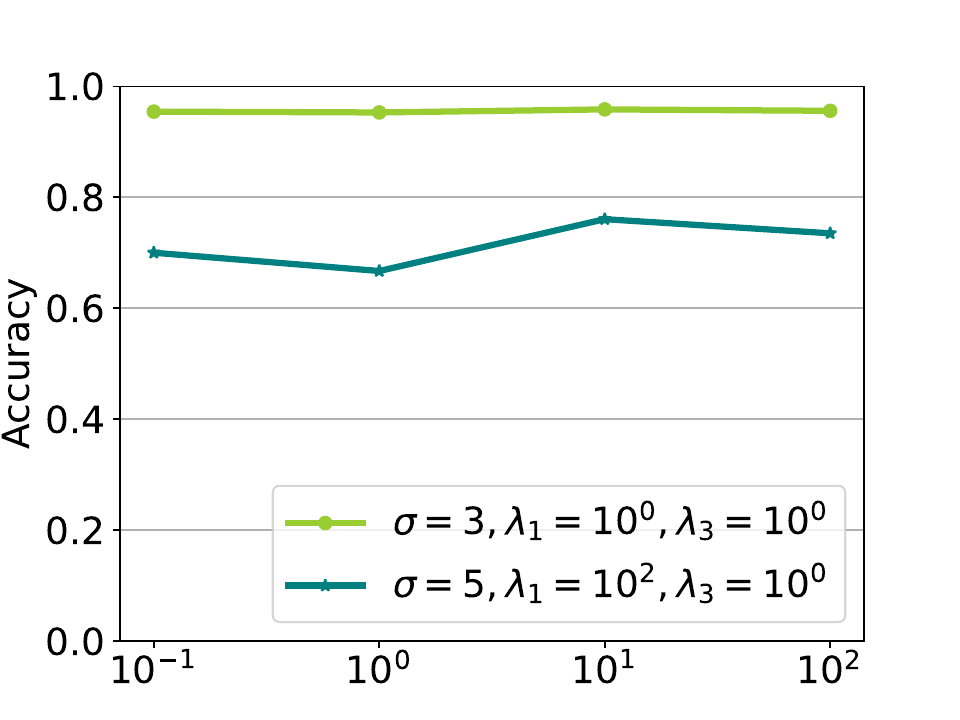}}
		\subfloat[\begin{scriptsize}
		    $\lambda_3$, Laplacian Noise
		\end{scriptsize}]{\includegraphics[width=0.3\linewidth]{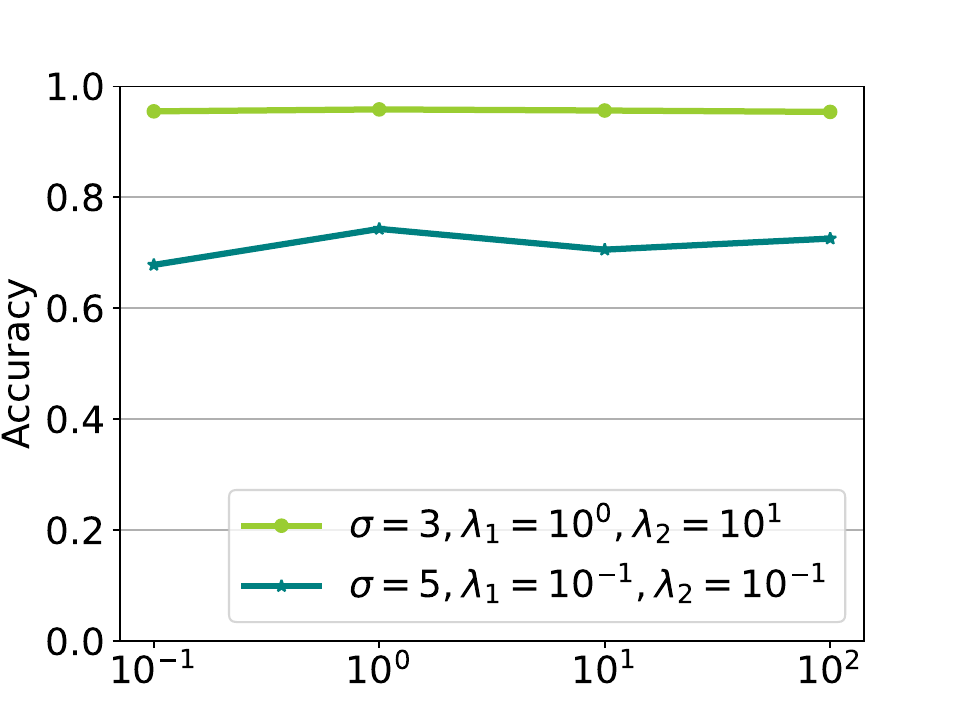}}
	\end{minipage}
	\caption{Parametric sensitivity of FLR on ``Aggre.''.}
\label{fig: the parametric on aggre with Gaussian and Laplacian}
\end{figure}

\subsection{Parametric Sensitivity}
Note that Eq.~\eqref{eq: the lagrangian function of feature and label noise} contains three trade-off parameters $\lambda_1$, $\lambda_2$, and $\lambda_3$ that need to be manually tuned. Therefore, we discuss whether the choices of them will significantly influence the performance of FLR. To this end, we examine the classification accuracy via changing one of $\lambda_1$, $\lambda_2$, and $\lambda_3$, and meanwhile fixing the others to the optimal constant values on ``Aggre.''. All experimental results are shown in Figure~\ref{fig: the parametric on aggre with Gaussian and Laplacian}.

We can observe that the performance of FLR is relatively stable in various noise scenarios. Therefore, the performance of our proposed FLR is actually insensitive to the choice of hyperparameters, and thus the parameters of our method can be easily tuned in practical uses. 

\begin{figure} [h!] 
	\begin{minipage}{1\linewidth}
        \captionsetup{font={scriptsize}} 
		\centering
		\subfloat[
		``Aggre.'']{\includegraphics[width=0.3\linewidth]{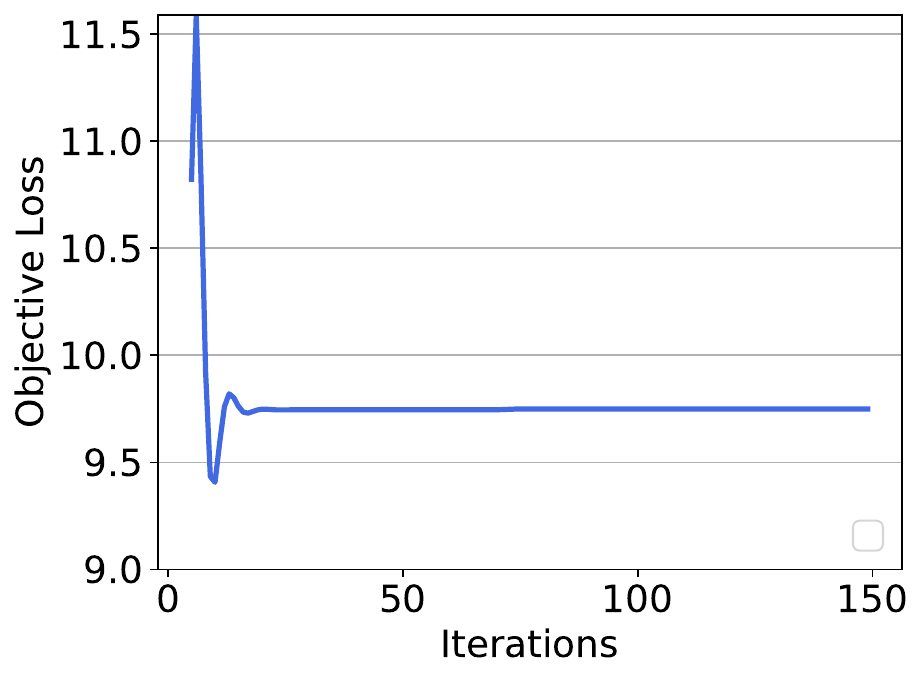}}
		\subfloat[\begin{scriptsize}
	    ``Random1''
		\end{scriptsize}]{\includegraphics[width=0.3\linewidth]{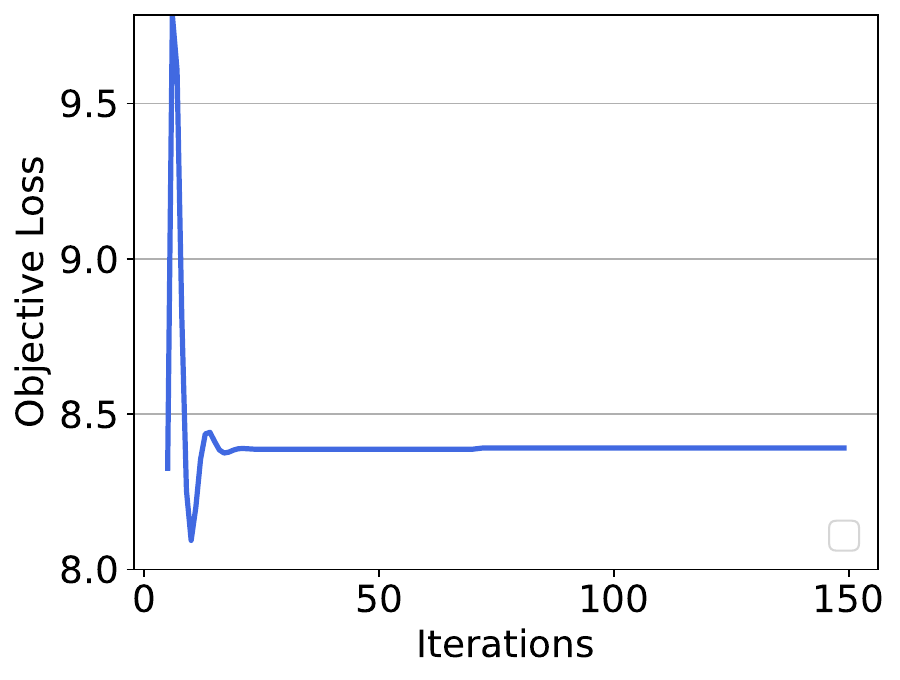}}
		\subfloat[\begin{scriptsize}
		  ``Worst''
		\end{scriptsize}]{\includegraphics[width=0.3\linewidth]{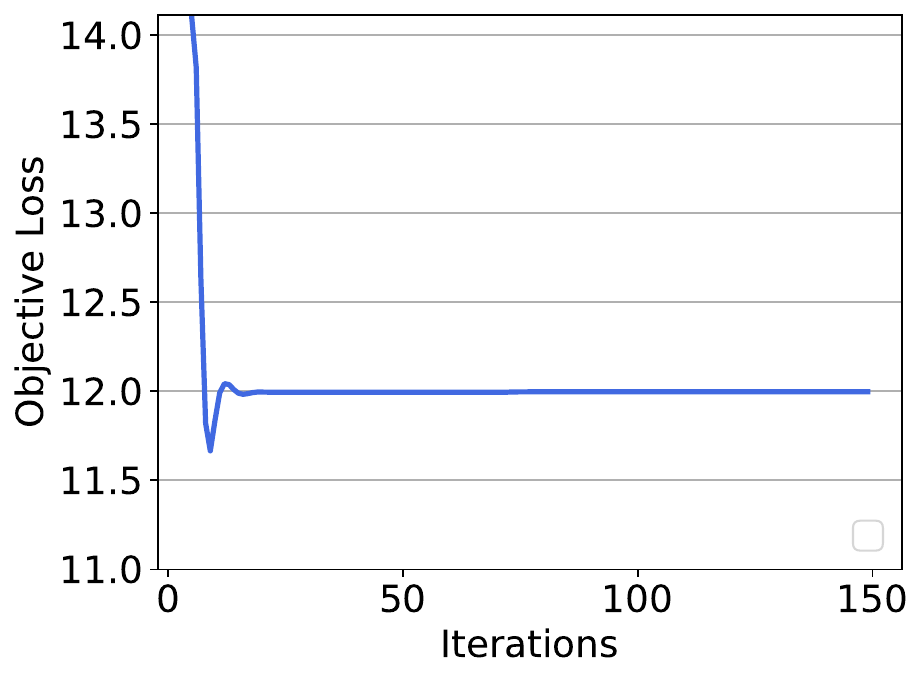}}
	\end{minipage}
	\caption{The illustration of convergence process of the ADMM method adopted by FLR on ``Aggre.'', ``Random1'' and ``Worst'' datasets.}
\label{fig: the loss convergence on Cifar10N}
\end{figure}

\subsection{Illustration of Convergence}
Note that we have theoretically proved that the optimization process in Algorithm~\ref{algorithm: the algorithm for feature and label noise} will converge to a stationary point. Here, we provide the convergence curves of FLR on the CIFAR-10N ``Aggre.'', ``Random1'' and ``Worst'' datasets. The curves shown in Figure~\ref{fig: the loss convergence on Cifar10N} justify our previous theoretical results and demonstrate that non-convex ADMM is effective~(and also efficient) in solving Eq.~\eqref{eq: equivalent form for feature and label correction with auxiliary variables}, where the loss function can always successfully converge to an accumulation point after sufficient iterations.

\section{Conclusion}

In this paper, we proposed a problem setting of hybrid noise, where we aim to learn a classifier from corrupted examples with noisy features and labels. The advantages of this paper lie in three aspects: 1) We built a novel unified robust learning paradigm to tackle the hybrid noise, where we jointly formulated the hybrid noise removal and classifier learning to a general data recovery objective; 2) We devised the optimization algorithm of non-convex ADMM for solving the proposed FLR, which is theoretically guaranteed to converge; 3) We also proved that the generalization error of FLR can be upper bounded, and it gradually shrinks even in the presence of the hybrid noise. Due to the aforementioned advantages of FLR, experimental results on several typical datasets demonstrated that FLR obtains higher classification accuracies than the existing feature noise learning and label noise learning methods in most cases. The learned classifier of FLR is limited by the fundamental assumption of linearity, which restricts its ability to effectively model and capture intricate, non-linear relationships inherent in many real-world datasets. Therefore, we intend to leverage the robust representational capabilities of deep learning to effectively manage hybrid noise within the aforementioned framework. In summary, this proposed FLR provides a heuristic framework to address hybrid noise by concurrently learning from both feature noise and label noise, which encourages us to address complex problems in a collaborative manner.

\bibliographystyle{spbasic}      % basic style, author-year citations
\bibliography{main}   % name your BibTeX data base

\begin{thebibliography}{44}
\providecommand{\natexlab}[1]{#1}
\providecommand{\url}[1]{{#1}}
\providecommand{\urlprefix}{URL }
\expandafter\ifx\csname urlstyle\endcsname\relax
  \providecommand{\doi}[1]{DOI~\discretionary{}{}{}#1}\else
  \providecommand{\doi}{DOI~\discretionary{}{}{}\begingroup \urlstyle{rm}\Url}\fi
\providecommand{\eprint}[2][]{\url{#2}}

\bibitem[{Cand{\`e}s et~al(2011)Cand{\`e}s, Li, Ma, and Wright}]{candes2011robust}
Cand{\`e}s EJ, Li X, Ma Y, Wright J (2011) Robust principal component analysis? Journal of the ACM 58(3):1--37

\bibitem[{Chen and Yang(2013)}]{chen2013robust}
Chen J, Yang J (2013) Robust subspace segmentation via low-rank representation. IEEE Transactions on Cybernetics 44(8):1432--1445

\bibitem[{Chen et~al(2017)Chen, Yang, Luo, Wei, Zhang, and Tai}]{chen2017low}
Chen S, Yang J, Luo L, Wei Y, Zhang K, Tai Y (2017) Low-rank latent pattern approximation with applications to robust image classification. IEEE Transactions on Image Processing 26(11):5519--5530

\bibitem[{Chen et~al(2019)Chen, Yang, Wei, Luo, Lu, and Gong}]{chen2019delta}
Chen S, Yang J, Wei Y, Luo L, Lu GF, Gong C (2019) $\delta$-norm-based robust regression with applications to image analysis. IEEE Transactions on Cybernetics 51(6):3371--3383

\bibitem[{Chiang et~al(2015)Chiang, Hsieh, and Dhillon}]{chiang2015matrix}
Chiang K, Hsieh C, Dhillon IS (2015) Matrix completion with noisy side information. In: Advances in Neural Information Processing Systems, pp 3447--3455

\bibitem[{Du et~al(2024)Du, Qiao, Yan, Zhang, and Zuo}]{du2024flexible}
Du J, Qiao X, Yan Z, Zhang H, Zuo W (2024) Flexible image denoising model with multi-layer conditional feature modulation. Pattern Recognition p 110372

\bibitem[{Gao(2008)}]{gao2008robust}
Gao J (2008) Robust l1 principal component analysis and its bayesian variational inference. Neural Computation 20(2):555--572

\bibitem[{Gao et~al(2016)Gao, Wang, Zhou et~al}]{gao2016risk}
Gao W, Wang L, Zhou ZH, et~al (2016) Risk minimization in the presence of label noise. In: Proceedings of the AAAI Conference on Artificial Intelligence, vol~30

\bibitem[{Ghosh et~al(2017)Ghosh, Kumar, and Sastry}]{ghosh2017robust}
Ghosh A, Kumar H, Sastry PS (2017) Robust loss functions under label noise for deep neural networks. In: Proceedings of the AAAI Conference on Artificial Intelligence, vol~31

\bibitem[{Gong et~al(2017)Gong, Zhang, Yang, and Tao}]{gong2017learning}
Gong C, Zhang H, Yang J, Tao D (2017) Learning with inadequate and incorrect supervision. In: 2017 IEEE International Conference on Data Mining, pp 889--894

\bibitem[{Gong et~al(2019)Gong, Shi, Liu, Zhang, Yang, and Tao}]{gong2019loss}
Gong C, Shi H, Liu T, Zhang C, Yang J, Tao D (2019) Loss decomposition and centroid estimation for positive and unlabeled learning. IEEE Transactions on Pattern Analysis and Machine Intelligence 43(3):918--932

\bibitem[{Grubbs(1973)}]{grubbs1973errors}
Grubbs FE (1973) Errors of measurement, precision, accuracy and the statistical comparison of measuring instruments. Technometrics 15(1):53--66

\bibitem[{Guo et~al(2021)Guo, Sun, Gao, Hu, and Yin}]{guo2021rank}
Guo J, Sun Y, Gao J, Hu Y, Yin B (2021) Rank consistency induced multiview subspace clustering via low-rank matrix factorization. IEEE Transactions on Neural Networks and Learning Systems 33(7):3157--3170

\bibitem[{Han et~al(2018{\natexlab{a}})Han, Yao, Niu, Zhou, Tsang, Zhang, and Sugiyama}]{han2018masking}
Han B, Yao J, Niu G, Zhou M, Tsang I, Zhang Y, Sugiyama M (2018{\natexlab{a}}) Masking: A new perspective of noisy supervision. In: Advances in Neural Information Processing Systems, vol~31

\bibitem[{Han et~al(2018{\natexlab{b}})Han, Yao, Yu, Niu, Xu, Hu, Tsang, and Sugiyama}]{han2018co}
Han B, Yao Q, Yu X, Niu G, Xu M, Hu W, Tsang I, Sugiyama M (2018{\natexlab{b}}) Co-teaching: Robust training of deep neural networks with extremely noisy labels. In: Advances in Neural Information Processing Systems, vol~31

\bibitem[{Hsieh et~al(2015)Hsieh, Natarajan, and Dhillon}]{hsieh2015pu}
Hsieh CJ, Natarajan N, Dhillon I (2015) Pu learning for matrix completion. In: Proceedings of International Conference on Machine Learning, pp 2445--2453

\bibitem[{Jiang et~al(2018)Jiang, Zhou, Leung, Li, and Fei-Fei}]{jiang2018mentornet}
Jiang L, Zhou Z, Leung T, Li LJ, Fei-Fei L (2018) Mentornet: Learning data-driven curriculum for very deep neural networks on corrupted labels. In: Proceedings of International Conference on Machine Learning, pp 2304--2313

\bibitem[{Ke et~al(2020)Ke, Gong, Liu, Zhao, Yang, and Tao}]{ke2020laplacian}
Ke J, Gong C, Liu T, Zhao L, Yang J, Tao D (2020) Laplacian welsch regularization for robust semisupervised learning. IEEE Transactions on Cybernetics 52(1):164--177

\bibitem[{Li et~al(2019)Li, Socher, and Hoi}]{li2019dividemix}
Li J, Socher R, Hoi SC (2019) Dividemix: Learning with noisy labels as semi-supervised learning. In: Proceedings of International Conference on Learning Representations

\bibitem[{Li et~al(2021)Li, Xiong, and Hoi}]{li2021learning}
Li J, Xiong C, Hoi SC (2021) Learning from noisy data with robust representation learning. In: Proceedings of the IEEE/CVF International Conference on Computer Vision, pp 9485--9494

\bibitem[{Lin et~al(2010)Lin, Chen, and Ma}]{lin2010augmented}
Lin Z, Chen M, Ma Y (2010) The augmented lagrange multiplier method for exact recovery of corrupted low-rank matrices. arXiv preprint arXiv:10095055

\bibitem[{Liu et~al(2010)Liu, Lin, and Yu}]{liu2010robust}
Liu G, Lin Z, Yu Y (2010) Robust subspace segmentation by low-rank representation. In: Proceedings of International Conference on Machine Learning, pp 663--670

\bibitem[{Luo et~al(2021)Luo, Han, and Gong}]{luo2021bi}
Luo Y, Han B, Gong C (2021) A bi-level formulation for label noise learning with spectral cluster discovery. In: Proceedings of International Conference on International Joint Conferences on Artificial Intelligence, pp 2605--2611

\bibitem[{Magoulas and Prentza(2001)}]{Magoulas2001}
Magoulas GD, Prentza A (2001) Machine learning in medical applications, Springer Berlin Heidelberg, Berlin, Heidelberg, pp 300--307

\bibitem[{Miranda et~al(2009)Miranda, Garcia, Carvalho, and Lorena}]{miranda2009use}
Miranda AL, Garcia LPF, Carvalho AC, Lorena AC (2009) Use of classification algorithms in noise detection and elimination. In: Hybrid Artificial Intelligence Systems, Springer, pp 417--424

\bibitem[{Muhlenbach et~al(2004)Muhlenbach, Lallich, and Zighed}]{muhlenbach2004identifying}
Muhlenbach F, Lallich S, Zighed DA (2004) Identifying and handling mislabelled instances. Journal of Intelligent Information Systems 22(1):89--109

\bibitem[{Patrini et~al(2016)Patrini, Nielsen, Nock, and Carioni}]{patrini2016lossFactorizationWeaklySupervisedLearning}
Patrini G, Nielsen F, Nock R, Carioni M (2016) Loss factorization, weakly supervised learning and label noise robustness. In: Proceedings of International Conference on Machine Learning, pp 708--717

\bibitem[{Royden and Fitzpatrick(2010)}]{royden2010real}
Royden H, Fitzpatrick PM (2010) Real analysis. China Machine Press

\bibitem[{Shi et~al(2014)Shi, Han, and Zheng}]{shi2014audio}
Shi Z, Han J, Zheng T (2014) Audio classification with low-rank matrix representation features. ACM Transactions on Intelligent Systems and Technology 5(1):1--17

\bibitem[{Tsouvalas et~al(2024)Tsouvalas, Saeed, Ozcelebi, and Meratnia}]{tsouvalas2024labeling}
Tsouvalas V, Saeed A, Ozcelebi T, Meratnia N (2024) Labeling chaos to learning harmony: Federated learning with noisy labels. ACM Transactions on Intelligent Systems and Technology 15(2):1--26

\bibitem[{Vincent et~al(2010)Vincent, Larochelle, Lajoie, Bengio, Manzagol, and Bottou}]{vincent2010stacked}
Vincent P, Larochelle H, Lajoie I, Bengio Y, Manzagol PA, Bottou L (2010) Stacked denoising autoencoders: Learning useful representations in a deep network with a local denoising criterion. Journal of Machine Learning Research 11(12)

\bibitem[{Wang et~al(2019)Wang, Ma, Chen, Luo, Yi, and Bailey}]{wang2019symmetric}
Wang Y, Ma X, Chen Z, Luo Y, Yi J, Bailey J (2019) Symmetric cross entropy for robust learning with noisy labels. In: Proceedings of the IEEE/CVF International Conference on Computer Vision, pp 322--330

\bibitem[{Wei et~al(2020)Wei, Feng, Chen, and An}]{wei2020combating}
Wei H, Feng L, Chen X, An B (2020) Combating noisy labels by agreement: A joint training method with co-regularization. In: Proceedings of the IEEE/CVF Conference on Computer Vision and Pattern Recognition, pp 13,726--13,735

\bibitem[{Wei et~al(2019)Wei, Gong, Chen, Liu, Yang, and Tao}]{wei2019harnessing}
Wei Y, Gong C, Chen S, Liu T, Yang J, Tao D (2019) Harnessing side information for classification under label noise. IEEE Transactions on Neural Networks and Learning Systems 31(9):3178--3192

\bibitem[{Wright et~al(2008)Wright, Yang, Ganesh, Sastry, and Ma}]{wright2008robust}
Wright J, Yang AY, Ganesh A, Sastry SS, Ma Y (2008) Robust face recognition via sparse representation. IEEE Transactions on Pattern Analysis and Machine Intelligence 31(2):210--227

\bibitem[{Xu et~al(2016)Xu, Tao, and Xu}]{xu2016robust}
Xu C, Tao D, Xu C (2016) Robust extreme multi-label learning. In: Proceedings of the ACM SIGKDD International Conference on Knowledge Discovery and Data Mining, pp 1275--1284

\bibitem[{Xu et~al(2019)Xu, Cao, Kong, and Wang}]{xu2019l_dmi}
Xu Y, Cao P, Kong Y, Wang Y (2019) L\_dmi: A novel information-theoretic loss function for training deep nets robust to label noise. In: Advances in Neural Information Processing Systems, vol~32

\bibitem[{Yang et~al(2016)Yang, Luo, Qian, Tai, Zhang, and Xu}]{yang2016nuclear}
Yang J, Luo L, Qian J, Tai Y, Zhang F, Xu Y (2016) Nuclear norm based matrix regression with applications to face recognition with occlusion and illumination changes. IEEE Transactions on Pattern Analysis and Machine Intelligence 39(1):156--171

\bibitem[{You et~al(2024)You, Liu, Han, and Xu}]{you2024beyond}
You Z, Liu D, Han B, Xu C (2024) Beyond pretrained features: Noisy image modeling provides adversarial defense. In: Advances in Neural Information Processing Systems, vol~36

\bibitem[{Zeng et~al(2022)Zeng, Yang, Chen, Yu, and Zhang}]{zeng2022clc}
Zeng B, Yang X, Chen Y, Yu H, Zhang Y (2022) Clc: A consensus-based label correction approach in federated learning. ACM Transactions on Intelligent Systems and Technology 13(5):1--23

\bibitem[{Zhang et~al(2018)Zhang, Yang, Shang, Gong, and Zhang}]{zhang2018lrr}
Zhang H, Yang J, Shang F, Gong C, Zhang Z (2018) Lrr for subspace segmentation via tractable schatten-$ p $ norm minimization and factorization. IEEE Transactions on Cybernetics 49(5):1722--1734

\bibitem[{Zhang et~al(2023)Zhang, Lv, Dai, Xin, and Dong}]{zhang2023noise}
Zhang J, Lv D, Dai Q, Xin F, Dong F (2023) Noise-aware local model training mechanism for federated learning. ACM Transactions on Intelligent Systems and Technology 14(4):1--22

\bibitem[{Zhang and Sabuncu(2018)}]{zhang2018generalized}
Zhang Z, Sabuncu M (2018) Generalized cross entropy loss for training deep neural networks with noisy labels. In: Advances in Neural Information Processing Systems, vol~31

\bibitem[{Zhu et~al(2022)Zhu, Wang, and Liu}]{zhu2022beyond}
Zhu Z, Wang J, Liu Y (2022) Beyond images: Label noise transition matrix estimation for tasks with lower-quality features. In: Proceedings of International Conference on Machine Learning, pp 27,633--27,653

\end{thebibliography}

\appendix
\section{Supplementary Material}
In this supplementary material, we provide the proofs for the theorems we proposed in the manuscript. The notations and numbering of equations follow those of the main paper.

\subsection{The Proof of Convergence Property (Theorem 1)}
Firstly, we provide the following lemma as a preliminary.
\begin{lemma}~\cite{lin2010augmented}
\label{lemma: the dual norm for M bounded}
Let $\mathcal{H}$ be a real Hilbert space endowed with an inner product 
$\langle \cdot, \cdot \rangle$, 
and a corresponding norm be 
$\Vert \cdot \Vert$. 
Considering 
$y \in \partial \Vert x \Vert$,
it holds that
$\Vert y \Vert^* = 1$,
if
$x \neq 0$,
and 
$\Vert y \Vert^* \leq 1$,
if 
$x = 0$,
where
$\partial f(\cdot)$
is the sub-gradient of 
$f(\cdot)$,
and 
$\Vert \cdot \Vert^*$ 
is the dual norm of 
$\Vert \cdot \Vert$.
\end{lemma}

Next, we provide the proof of convergence property exhaustively.

\begin{proof}
    \begin{enumerate}
        \item Here, we will prove that the generated sequences
        $\{\{\bm{M}_{i,t}\}_{i=1}^5\}$,
        $\{\bm{X}_t\}$, 
        $\{\bm{Z}_t\}$, 
        $\{\bm{B}_t\}$,
        $\{\bm{J}_t\}$, 
        $\{\bm{K}_t\}$, 
        $\{\bm{E}_{f, t}\}$, 
        and
        $\{\bm{E}_{l, t}\}$
        are all bounded.\\
        According to the updating rule of $\bm{B}$ in Eq.~\eqref{eq: the solution to subproblem B}, $\bm{B} \in \left[0, 1\right]^{n \times c}$ always holds, and thus the sequence $\{\bm{B}_t\}$ is bounded.\\
        As stated in the definition of the augmented Lagrangian function $\mathcal{L}$, {\em{i.e.}}, Eq.~\eqref{eq: the lagrangian function of feature and label noise}, and the first-order optimality condition of $\bm{E}_{f, t + 1}$ and $\bm{E}_{l, t + 1}$, we have 
            \begin{equation}
            \begin{aligned}
                \bm{0} 
                &\in \partial \mathcal{L}_{\bm{E}_f}\left(\bm{X}_{t + 1},  \bm{Z}_{t + 1},  \bm{B}_{t + 1},  \bm{J}_{t + 1},  \bm{K}_{t + 1},  \bm{E}_{f, t + 1},  \bm{E}_{l, t},  \{\bm{M}_{i, t}\}_{i = 1}^5\right)\\
                & = \partial(\lambda_2 \Vert \bm{E}_{f, t + 1}\Vert_1) - \bm{M}_{1, t}
                - \mu_t(\bm{\tilde{X}} - \bm{X}_{t + 1} - \bm{E}_{f, t + 1}),
            \end{aligned}
        \end{equation}
        as \begin{equation}
            \bm{M}_{1, t+ 1} = \bm{M}_{1, t} + \mu_t(\bm{\tilde{X}} - \bm{X}_{t + 1} - \bm{E}_{f, t + 1}),
        \end{equation}
        namely
        \begin{equation}
            \begin{aligned}
                \bm{0} \in \partial(\lambda_2 \Vert \bm{E}_{f, t + 1}\Vert_1) - \bm{M}_{1, t + 1}.
            \end{aligned}
        \end{equation}
        So we have that 
        \begin{equation}
            \begin{aligned}
               \bm{M}_{1, t + 1} \in \partial(\lambda_2 \Vert \bm{E}_{f, t + 1}\Vert_1).
            \end{aligned}
        \end{equation}
        Then by Lemma~\ref{lemma: the dual norm for M bounded}, it holds that
        \begin{equation}
            \begin{aligned}
                \Vert\bm{M_{1, t+1}}\Vert_1^* \leq 1.
            \end{aligned}
        \end{equation}
        Therefore, the sequence $\{\bm{M}_{1,t}\}$ is bounded. Similarly, the sequence $\{\bm{M}_{2,t}\}$ can be proved to be bounded by calculating the partial derivative of $\mathcal{L}$ {\em{w.r.t.}}\ $\bm{E}_{l,t + 1}$.\\
        
        Then by the optimality of $\bm{Z}_{t + 1}$, 
        \begin{equation}
        \label{eq: the subgradient of L_z}
            \begin{aligned}
                \bm{0} \in \partial\mathcal{L}_{\bm{Z}}\left(\bm{X}_{t + 1},  \bm{Z}_{t + 1},  \bm{B}_{t},  \bm{J}_{t},  \bm{K}_{t},  \bm{E}_{f},  \bm{E}_{l},  \{\bm{M}_{i, t}\}_{i = 1}^5\right),
            \end{aligned}
        \end{equation}
        namely
        \begin{equation}
            \begin{aligned}
                \bm{0} 
                &\in \partial\left(\lambda_1 \Vert\bm{Z}_{t + 1}\Vert_*\right) + \bm{M}_{3, t} + \mu_t(\bm{Z}_{t+1} - \bm{J}_{t})\\
                & =\partial\left(\lambda_1 \Vert\bm{Z}_{t + 1}\Vert_*\right) + \bm{M}_{3, t}  +
                \mu_t(\bm{Z}_{t+1} - \bm{J}_{t + 1}) + \mu_t(\bm{J}_{t+1} - \bm{J}_{t}).
            \end{aligned}
        \end{equation}
        
        As
        \begin{equation}
            \bm{M}_{3, t+ 1} = \bm{M}_{3, t} + \mu_t(\bm{Z}_{t + 1} - \bm{J}_{t + 1}),
        \end{equation}
        we see that
        \begin{equation}
        \label{eq: the dual norm of M3 for feature and label noise}
            \begin{aligned}
                \Vert-\bm{M}_{3, t + 1} - \mu_t (\bm{J}_{t+1} - \bm{J}_{t})\Vert_*^* \leq 1.
            \end{aligned}
        \end{equation}
        Combining Eq.~\eqref{eq: the dual norm of M3 for feature and label noise} with
        \begin{equation}
            \lim\limits_{t \rightarrow +\infty} 
        \mu_t(\bm{J}_{t + 1} - \bm{J}_t) = 0,
        \end{equation}
        we conclude that the sequence $\{\bm{M}_{3, t}\}$ is bounded.

        In the same way, considering
        \begin{equation}
        \label{eq: the subgradient of L_x}
            \begin{aligned}
                \bm{0} 
                & \in \partial\mathcal{L}_{\bm{X}}\left(\bm{X}_{t + 1},  \bm{Z}_{t},  \bm{B}_{t},  \bm{J}_{t},  \bm{K}_{t},  \bm{E}_{f},  \bm{E}_{l},  \{\bm{M}_{i, t}\}_{i = 1}^5\right)\\
                & \in \partial\Vert\bm{X}_{t + 1}\Vert_* - \bm{M}_{1, t} + \bm{M}_{5, t} 
                 + \mu_t\left(-\left(\bm{\tilde{X}} -  \bm{X}_{t + 1} - \bm{E}_{f, t}\right) + \left(\bm{X}_{t + 1} - \bm{K}_t\right)\right)\\
                & \in \partial\Vert\bm{X}_{t + 1}\Vert_* - \bm{M}_{1, t + 1} + \bm{M}_{5, t + 1}
                 - \mu_t(\bm{E}_{f, t + 1} - \bm{E}_{f, t}) + \mu_t (\bm{K}_{t + 1 } - \bm{K}_t),
            \end{aligned}
        \end{equation}
        we have 
            \begin{equation}
            \label{eq: the dual norm of M5 for feature and label noise}
            \begin{aligned}
                \Vert\bm{M}_{1, t+1} -  \bm{M}_{5, t+1} + \mu_t(\bm{E}_{f, t + 1} - \bm{E}_{f, t}) - \mu_t (\bm{K}_{t + 1 } - \bm{K}_t)\Vert_*^* \leq 1.
            \end{aligned}
        \end{equation}
        Also, we combine Eq.~\eqref{eq: the dual norm of M5 for feature and label noise} with 
        \begin{equation}
            \begin{aligned}
                &\lim\limits_{t \rightarrow +\infty} 
                \mu_t(\bm{K}_{t + 1} - \bm{K}_t) = 0, \\ 
                &\lim\limits_{t \rightarrow +\infty}
                \mu_t(\bm{E}_{f, t + 1} - \bm{E}_{f, t}) = 0,
            \end{aligned}    
        \end{equation}
        and the boundedness of sequence $\{\bm{M}_{1,t}\}$, the boundedness of sequence $\{\bm{M}_{5, t}\}$ can be proved.\\
        Next, by the optimality of $\bm{B}_{t+1}$,
        we have 
        \begin{equation}
            \begin{aligned}
                \bm{0} 
                & = - \bm{M}_{2,t} + \bm{M}_{4, t} - \mu_t\left(\bm{\tilde{Y}} - \bm{B}_{t + 1} - \bm{E}_{l, t} \right)
                 + \mu_t\left(\bm{B}_{t+1} -\bm{K}_t\bm{J}_t\right),
            \end{aligned}
        \end{equation}
        which leads to      
        \begin{equation}
            \begin{aligned}
                \bm{M}_{2,t + 1} =  
                \bm{M}_{4, t + 1} - \mu_t(\bm{E}_{l,t+1} - \bm{E}_{l,t})
                + \mu_t(\bm{K}_{t + 1}\bm{J}_{t + 1}-\bm{K}_t\bm{J}_t) .
            \end{aligned}
        \end{equation}
        Based on the assumptions of 
        $\lim\limits_{t \rightarrow +\infty} 
        \mu_t(\bm{K}_{t + 1} - \bm{K}_t) = 0$, 
        and
        $\lim\limits_{t \rightarrow +\infty} 
        \mu_t(\bm{J}_{t + 1} - \bm{J}_t) = 0$, 
        we have $\lim\limits_{t \rightarrow +\infty} 
        \mu_t(\bm{K}_{t + 1}\bm{J}_t - \bm{K}_t\bm{J}_t) = 0$. Also, we have proved the boundedness of $\bm{M}_{2,t+1}$ and assume $\lim\limits_{t \rightarrow +\infty}
        \mu_t(\bm{E}_{l, t + 1} - \bm{E}_{l, t}) = 0$, so it is obviously the sequence $\bm{M}_{4,t+1}$ is bounded.   

        Thus far, we have proved that these sequences $\{\bm{M}_{i, t}\}_{i = 1}^5$ are all bounded.

        In addition, we here derive the following chain of equations:
        \begin{equation}
            \begin{aligned}
                \mathcal{L}
                & \left(\bm{X}_{t},  \bm{Z}_{t},  \bm{B}_{t},  \bm{J}_{t},  \bm{K}_{t},  \bm{E}_{f},  \bm{E}_{l},  \{\bm{M}_{i, t}\}_{i = 1}^5\right)\\
                & = \Vert \bm{X}_t\Vert_* + 
	        \lambda_1 \Vert\bm{Z}_t\Vert_* + 
	        \lambda_2 \Vert \bm{E}_{f,t}\Vert_1 + 
	        \lambda_3 \Vert\bm{E}_{l,t}\Vert_{2,1} \\
                & + \Phi(\bm{M}_{1,t}, \bm{\tilde{X}} - \bm{X}_t - \bm{E}_{f,t}) 
                  + \Phi(\bm{M}_{2,t}, \bm{\tilde{Y}} - \bm{B}_t - \bm{E}_{l,t}) \\
                & + \Phi(\bm{M}_{3,t}, \bm{Z}_t - \bm{J}_t) 
                  + \Phi(\bm{M}_{4,t}, \bm{B}_t - \bm{K}_t\bm{J}_t)
                  + \Phi(\bm{M}_{5,t}, \bm{X}_t - \bm{K}_t),
            \end{aligned}
        \end{equation}
        where $\Phi(\bm{A}, \bm{B}) = \frac{\mu}{2}\Vert\bm{B}\Vert_F^2 + \langle\bm{A}, \bm{B}\rangle$;
        
        \begin{equation}
            \begin{aligned}
                \mathcal{L}
                & \left(\bm{X}_{t},  \bm{Z}_{t},  \bm{B}_{t},  \bm{J}_{t},  \bm{K}_{t},  \bm{E}_{f},  \bm{E}_{l},  \{\bm{M}_{i, t-1}\}_{i = 1}^5\right)\\
                & = \Vert \bm{X}_t\Vert_* + 
	        \lambda_1 \Vert\bm{Z}_t\Vert_* + 
	        \lambda_2 \Vert \bm{E}_{f,t}\Vert_1 + 
	        \lambda_3 \Vert\bm{E}_{l,t}\Vert_{2,1} \\
                & + \Phi(\bm{M}_{1,t-1}, \bm{\tilde{X}} - \bm{X}_t - \bm{E}_{f,t}) 
                  + \Phi(\bm{M}_{2,t-1}, \bm{\tilde{Y}} - \bm{B}_t - \bm{E}_{l,t}) \\
                & + \Phi(\bm{M}_{3,t-1}, \bm{Z}_t - \bm{J}_t) 
                  + \Phi(\bm{M}_{4,t-1}, \bm{B}_t - \bm{K}_t\bm{J}_t)
                  + \Phi(\bm{M}_{5,t-1}, \bm{X}_t - \bm{K}_t),
            \end{aligned}
        \end{equation}
        namely
        \begin{equation}
        \label{eq: the relation of M_t and M_t -1}
            \begin{aligned}
                \mathcal{L}
                & \left(\bm{X}_{t},  \bm{Z}_{t},  \bm{B}_{t},  \bm{J}_{t},  \bm{K}_{t},  \bm{E}_{f, t},  \bm{E}_{l, t},  \{\bm{M}_{i, t}\}_{i = 1}^5\right)\\
                & = \mathcal{L} \left(\bm{X}_{t},  \bm{Z}_{t},  \bm{B}_{t},  \bm{J}_{t},  \bm{K}_{t},  \bm{E}_{f,t},  \bm{E}_{l,t},  \{\bm{M}_{i, t-1}\}_{i = 1}^5\right)\\
                & + \langle \bm{M}_{1,t} - \bm{M}_{1,t-1}, \bm{\tilde{X}} - \bm{X}_t - \bm{E}_{f,t}\rangle\\
                & + \langle\bm{M}_{2,t} - \bm{M}_{2,t-1}, \bm{\tilde{Y}} - \bm{B}_t - \bm{E}_{l,t}\rangle 
                 + \langle\bm{M}_{3,t} - \bm{M}_{3,t-1}, \bm{Z}_t - \bm{J}_t\rangle \\
                & + \langle\bm{M}_{4,t} - \bm{M}_{4,t-1}, \bm{B}_t - \bm{K}_t\bm{J}_t\rangle 
                 + \langle\bm{M}_{5,t} - \bm{M}_{5,t-1}, \bm{X}_t - \bm{K}_t\rangle \\
                & + \frac{\mu_t - \mu_{t -1}}{2}(
                \Vert\bm{\tilde{X}} - \bm{X}_t - \bm{E}_{f,t}\Vert_F^2 + \Vert\bm{\tilde{Y}} - \bm{B}_t - \bm{E}_{l,t}\Vert_F^2 \\
                 & + \Vert\bm{Z}_t - \bm{J}_t\Vert_F^2  + \Vert\bm{B}_t - \bm{K}_t\bm{J}_t\Vert_F^2 + \Vert\bm{X}_t - \bm{K}_t\Vert_F^2
                )
                \\
                & = \mathcal{L} \left(\bm{X}_{t},  \bm{Z}_{t},  \bm{B}_{t},  \bm{J}_{t},  \bm{K}_{t},  \bm{E}_{f,t},  \bm{E}_{l,t},  \{\bm{M}_{i, t-1}\}_{i = 1}^5\right)\\
                & + \frac{\mu_t + \mu_{t - 1}}{2\mu_{t-1}^2}
                \sum\limits_{i = 1}^5\Vert\bm{M}_{i, t} - \bm{M}_{i, t-1} \Vert_F^2.
            \end{aligned}
        \end{equation}
        
        Since Algorithm~\ref{algorithm: the algorithm for feature and label noise} optimizes variables $\bm{X}$, $\bm{Z}$, $\bm{B}$, $\bm{J}$, $\bm{K}$, $\bm{E}_f$ and $\bm{E}_l$ alternatively, according to the optimality of them and Eq.~\eqref{eq: the relation of M_t and M_t -1}, we have
        \begin{equation}
        \label{eq: the relation between t and t-1 of lagrangian}
            \begin{aligned}
                \mathcal{L}&\left(\bm{X}_{t+1},  \bm{Z}_{t+1},  \bm{B}_{t+1},  \bm{J}_{t+1},  \bm{K}_{t+1},  \bm{E}_{f, t+1},  \bm{E}_{l, t+1},  \{\bm{M}_{i, t}\}_{i = 1}^5\right) \\
                & \leq \mathcal{L} \left(\bm{X}_{t},  \bm{Z}_{t},  \bm{B}_{t},  \bm{J}_{t},  \bm{K}_{t},  \bm{E}_{f,t},  \bm{E}_{l,t},  \{\bm{M}_{i, t}\}_{i = 1}^5\right)\\
                &=  \mathcal{L} \left(\bm{X}_{t},  \bm{Z}_{t},  \bm{B}_{t},  \bm{J}_{t},  \bm{K}_{t},  \bm{E}_{f,t},  \bm{E}_{l,t},  \{\bm{M}_{i, t-1}\}_{i = 1}^5\right)\\
                 &+ \frac{\mu_t + \mu_{t - 1}}{2\mu_{t-1}^2}
                \sum\limits_{i = 1}^5\Vert\bm{M}_{i, t} - \bm{M}_{i, t-1} \Vert_F^2,
            \end{aligned}
        \end{equation}
        by setting $t = n$, and taking the relationship in Eq.~\eqref{eq: the relation between t and t-1 of lagrangian} into consideration, it holds that 
        \begin{equation}
        \label{eq: the summary of each term of lagrangian for feature and label noise}
            \begin{aligned}
                \mathcal{L}&\left(\bm{X}_{n+1},  \bm{Z}_{n+1},  \bm{B}_{n+1},  \bm{J}_{n+1},  \bm{K}_{n+1},  \bm{E}_{f, n+1},  \bm{E}_{l, n+1},  \{\bm{M}_{i, n}\}_{i = 1}^5\right) \\
                &\leq  \mathcal{L} \left(\bm{X}_{1},  \bm{Z}_{1},  \bm{B}_{1},  \bm{J}_{1},  \bm{K}_{1},  \bm{E}_{f,1},  \bm{E}_{l,1},  \{\bm{M}_{i,0}\}_{i = 1}^5\right)\\
                &+ \sum\limits_{t = 1}^n\left(\frac{\mu_t + \mu_{t - 1}}{2\mu_{t-1}^2}
                \sum\limits_{i = 1}^5\Vert\bm{M}_{i, t} - \bm{M}_{i, t-1} \Vert_F^2\right).
            \end{aligned}
        \end{equation}
        Note that we have proved the sequences $\{\bm{M}_{i,t}\}_{i=1}^5$ are bounded, so there exists one constant $\mathcal{C}$, and then $\sum\limits_{i = 1}^5\Vert\bm{M}_{i, t} - \bm{M}_{i, t-1} \Vert_F^2 \leq \mathcal{C}$.

        Therefore, 
        \begin{equation}
            \begin{aligned}
             &\sum\limits_{t = 1}^n\left(\frac{\mu_t + \mu_{t - 1}}{2\mu_{t-1}^2}
                \sum\limits_{i = 1}^5\Vert\bm{M}_{i, t} - \bm{M}_{i, t-1} \Vert_F^2\right)\\
             &\leq  \sum\limits_{t = 1}^n\left(\frac{\mu_t + \mu_{t - 1}}{2\mu_{t-1}^2}
                \mathcal{C}\right)\\
             &= \mathcal{C}\sum\limits_{t = 1}^n\frac{\mu_0\rho^t + \mu_0\rho^{t - 1}}{2(\mu_0\rho^{t-1})^2}\\
             &= \mathcal{C}\frac{\rho+1}{2\mu_0}\sum\limits_{t = 1}^n\frac{1}{\rho^{t-1}}
        \leq \mathcal{C}\frac{(\rho+1)\rho}{2\mu_0(\rho - 1)}(1 - \frac{1}{\rho^n})
        < + \infty,
            \end{aligned}
        \end{equation}
        in which $\rho^t$ denotes the exponential power calculation and it holds that $\mu_t = \mu_0\rho^t$.
        
        Now we can see that both sides of Eq.~\eqref{eq: the summary of each term of lagrangian for feature and label noise} are bounded. Based on this conclusion and the boundedness of $\{\bm{M}_{i, t}\}_{i = 1}^5$, each term of the following equation is bounded, namely
        \begin{equation}
        \label{eq: the final boundedness of each variable}
            \begin{aligned}
                \mathcal{L}&\left(\bm{X}_{t+1},  \bm{Z}_{t+1},  \bm{B}_{t+1},  \bm{J}_{t+1},  \bm{K}_{t+1},  \bm{E}_{f, t+1},  \bm{E}_{l, t+1},  \{\bm{M}_{i, t}\}_{i = 1}^5\right)
                 + \frac{1}{2\mu_t}\sum\limits_{i = 1}^5 \Vert\bm{M}_{i,t}\Vert_F^2\\
                & = \Vert\bm{X}_{t + 1}\Vert_* + \lambda_1 \Vert\bm{Z}_{t +1}\Vert_* + \lambda_2\Vert\bm{E}_{f, t+1}\Vert_1 +\lambda_3\Vert\bm{E}_{l, t+1}\Vert_{2,1}\\
                & + \Phi(\bm{M}_{1,t}, \bm{\tilde{X}} - \bm{X}_t - \bm{E}_{f,t}) + \frac{1}{2\mu_t}\Vert\bm{M}_{1,t}\Vert_F^2\\
                & + \Phi(\bm{M}_{2,t}, \bm{\tilde{Y}} - \bm{B}_t - \bm{E}_{l,t}) + \frac{1}{2\mu_t}\Vert\bm{M}_{2,t}\Vert_F^2\\
                & + \Phi(\bm{M}_{3,t}, \bm{Z}_t - \bm{J}_t) + \frac{1}{2\mu_t}\Vert\bm{M}_{3,t}\Vert_F^2\\
                & + \Phi(\bm{M}_{4,t}, \bm{B}_t - \bm{K}_t\bm{J}_t)+ \frac{1}{2\mu_t}\Vert\bm{M}_{4,t}\Vert_F^2\\
                & + \Phi(\bm{M}_{5,t}, \bm{X}_t - \bm{K}_t) + \frac{1}{2\mu_t}\Vert\bm{M}_{5,t}\Vert_F^2.
                \end{aligned}
                \end{equation}
                Then,
                \begin{equation}
                    \begin{aligned}
                    \mathcal{L}&\left(\bm{X}_{t+1},  \bm{Z}_{t+1},  \bm{B}_{t+1},  \bm{J}_{t+1},  \bm{K}_{t+1},  \bm{E}_{f, t+1},  \bm{E}_{l, t+1},  \{\bm{M}_{i, t}\}_{i = 1}^5\right)
                 + \frac{1}{2\mu_t}\sum\limits_{i = 1}^5 \Vert\bm{M}_{i,t}\Vert_F^2\\
                & = \Vert\bm{X}_{t + 1}\Vert_* + \lambda_1 \Vert\bm{Z}_{t +1}\Vert_* + \lambda_2\Vert\bm{E}_{f, t+1}\Vert_1 +\lambda_3\Vert\bm{E}_{l, t+1}\Vert_{2,1} \\
                & + \frac{\mu_t}{2}\left(\Vert \bm{\tilde{X}} - \bm{X}_{t + 1} - \bm{E}_{f,t+1} + \frac{\bm{M}_{1,t}}{\mu_t}\Vert_F^2\right)
                 + \frac{\mu_t}{2}\left(\Vert \bm{\tilde{Y}} - \bm{B}_{t + 1} - \bm{E}_{l,t+1} + \frac{\bm{M}_{2,t}}{\mu_t}\Vert_F^2\right)\\
                & + \frac{\mu_t}{2}\left(\Vert \bm{Z}_{t + 1} - \bm{J}_{t + 1} + \frac{\bm{M}_{3,t}}{\mu_t}\Vert_F^2\right)
                 + \frac{\mu_t}{2}\left(\Vert \bm{B}_{t + 1} - \bm{K}_{t + 1}\bm{J}_{t + 1} + \frac{\bm{M}_{4,t}}{\mu_t}\Vert_F^2\right)\\
                & + \frac{\mu_t}{2}\left(\Vert \bm{X}_{t + 1} - \bm{K}_{t + 1} + \frac{\bm{M}_{5,t}}{\mu_t}\Vert_F^2\right).
            \end{aligned}
        \end{equation}
        Specifically, it is obvious that $\{\bm{X}_t\}$, $\{\bm{Z}_t\}$, $\{\bm{E}_{f,t}\}$, and $\{\bm{E}_{l,t}\}$ are bounded. From the ninth term of Eq.~\eqref{eq: the final boundedness of each variable}, we can obtain the boundedness of $\bm{K}_{t}$. Moreover, from the seventh term of Eq.~\eqref{eq: the final boundedness of each variable}, $\bm{J}_{t}$ is bounded.

        Up to now, we have proved that the sequence $\{\Gamma_t = (\bm{X}_t, \bm{Z}_t, \bm{B}_t, \bm{J}_t, \bm{K}_t, \bm{E}_{l, t}, \bm{E}_{f, t}, \{\bm{M}_{i,t}\}_{i=1}^5 )\}_{t = 1}^\infty$ is bounded.

        \item According to the analysis mentioned above, and by the Bolzano-Weierstrass theorem~\cite{royden2010real}, the sequence $\{\Gamma_t\}$ must have at least one accumulation point. Then $\{\Gamma^* = (\bm{X}^*, \bm{Z}^*, \bm{B}^*, \bm{J}^*, \bm{K}^*, \bm{E}_l^*, \bm{E}_f^*, \\ \{\bm{M}_{i}^*\}_{i=1}^5 )\}$ denotes the accumulation point. Without loss of generality, we assume that the sequence $\Gamma_t$ converges to $\Gamma^*$.

        As 
        \begin{equation}
            \begin{aligned}
                &\bm{M}_{1,t+1} = \bm{M}_{1,t} + \mu_t\left(\bm{\tilde{X}} - \bm{X}_{t + 1} - \bm{E}_{f, t+1}\right),\\
		    &\bm{M}_{2,t+1} = \bm{M}_{2,t} + \mu_t\left(\bm{\tilde{Y}} - \bm{B}_{t+1} - \bm{E}_{l,t+1}\right),\\
		    &\bm{M}_{3,t+1} = \bm{M}_{3,t} + \mu_t\left(\bm{Z}_{t+1}-\bm{J}_{t+1}\right),\\
		    &\bm{M}_{4,t+1} = \bm{M}_{4,t} + \mu_t\left(\bm{B}_{t+1} - \bm{K}_{t+1}\bm{J}_{t+1}\right),\\
		    &\bm{M}_{5,t+1} = \bm{M}_{5,t} + \mu_t\left(\bm{X}_{t+1} - \bm{K}_{t+1}\right),
            \end{aligned}
        \end{equation}
    we have 
    \begin{equation}
            \begin{aligned}
                & \bm{\tilde{X}} - \bm{X}_{t+1} - \bm{E}_{f, t+1} = \frac{1}{\mu_t}\left(\bm{M}_{1,t+1} - \bm{M}_{1,t}\right),\\
		    &\bm{\tilde{Y}} - \bm{B}_{t+1} - \bm{E}_{l,t+1} = \frac{1}{\mu_t}\left(\bm{M}_{2,t+1} - \bm{M}_{2,t}\right),\\
		    &\bm{Z}_{t+1}-\bm{J}_{t+1} = \frac{1}{\mu_t}\left(\bm{M}_{3,t+1} - \bm{M}_{3,t}\right),\\
		    &\bm{B}_{t+1} - \bm{K}_{t+1}\bm{J}_{t+1} = \frac{1}{\mu_t}\left(\bm{M}_{4,t+1} - \bm{M}_{4,t}\right),\\
		    &\bm{X}_{t+1} - \bm{K}_{t+1} = \frac{1}{\mu_t}\left(\bm{M}_{5,t+1} - \bm{M}_{5,t}\right).
            \end{aligned}
        \end{equation}
    Because of the boundedness of $\{\bm{M}_{i,t}\}_{i = 1}^5$ and when $t \rightarrow + \infty$, it holds that 
    \begin{equation}
        \begin{aligned}
            & \bm{\tilde{X}} - \bm{X}_{t+1} - \bm{E}_{f, t+1} \rightarrow 0,\\
		&\bm{\tilde{Y}} - \bm{B}_{t+1} - \bm{E}_{l,t+1} \rightarrow 0,\\
		&\bm{Z}_{t+1}-\bm{J}_{t+1} \rightarrow 0,\\
		&\bm{B}_{t+1} - \bm{K}_{t+1}\bm{J}_{t+1} \rightarrow 0,\\
		&\bm{X}_{t+1} - \bm{K}_{t+1} \rightarrow 0,
        \end{aligned}
    \end{equation}
    and then we have
    \begin{equation}
        \begin{aligned}
            & \bm{\tilde{X}} - \bm{X}^* - \bm{E}_f^* \rightarrow 0,\\
		&\bm{\tilde{Y}} - \bm{B}^* - \bm{E}_l^* \rightarrow 0,\\
		&\bm{Z}^*-\bm{J}^* \rightarrow 0,\\
		&\bm{B}^* - \bm{K}^*\bm{J}^* \rightarrow 0,\\
		&\bm{X}^* - \bm{K}^* \rightarrow 0.
        \end{aligned}
    \end{equation}
    
    In addition, based on the updating rule of $\bm{B}$, $\bm{B} \in [0, 1]$ always holds, and thus $\bm{B}^* \in [0, 1]$. Hence, the primal feasibility conditions of the optimization problem are satisfied by $\Gamma^*$.

    Moreover, we will prove that the first-order stationary condition holds when $\Gamma = \Gamma^*$.
    According to the updating rules for $\bm{X}_{t + 1}$, $\bm{M}_{1, t+1}$ and $\bm{M}_{5, t+1}$, {\em{i.e.}}, Eq.~\eqref{eq: the subgradient of L_x}, we have 
    \begin{equation}
        \begin{aligned}
           0 & \in \partial\Vert\bm{X}_{t + 1}\Vert_* - \bm{M}_{1, t + 1} + \bm{M}_{5, t + 1}\\
             &\ \ \ - \mu_t(\bm{E}_{f, t + 1} - \bm{E}_{f, t}) + \mu_t (\bm{K}_{t + 1 } - \bm{K}_t).
        \end{aligned}
    \end{equation}
    Under the assumptions that 
    \begin{equation}
        \begin{aligned}
        &\lim\limits_{t \rightarrow +\infty} 
        \mu_t(\bm{K}_{t + 1} - \bm{K}_t) = 0, \\ 
        &\lim\limits_{t \rightarrow +\infty}
        \mu_t(\bm{E}_{f, t + 1} - \bm{E}_{f, t}) = 0,
        \end{aligned}
    \end{equation}
    and let $t \rightarrow + \infty$, we have
    \begin{equation}
        \begin{aligned}
            \bm{0} \in \partial\Vert\bm{X}^*\Vert_* - \bm{M}_1^* + \bm{M}_5^*,
        \end{aligned}
    \end{equation}
    which leads to
    \begin{equation}
        \begin{aligned}
            \partial\mathcal{L}_{\bm{X}}\left(   \bm{X},  \bm{Z},  \bm{B},  \bm{J},  \bm{K},  \bm{E}_f,  \bm{E}_l,  \bm{M}_{i = 1}^5 \right)|_{(\bm{X} = \bm{X}^*)} = 0.
        \end{aligned}
    \end{equation}

    By the rule of $\bm{Z}_{t + 1}$, {\em{i.e.}}, Eq.~\eqref{eq: the subgradient of L_z} and $\lim\limits_{t \rightarrow +\infty} \mu_t(\bm{J}_{t + 1} - \bm{J}_t) = 0$,
    we have 
    \begin{equation}
        \begin{aligned}
            &\bm{0} \in \partial\left(\lambda_1 \Vert\bm{Z}_{t + 1}\Vert_*\right) + \bm{M}_{3, t + 1} + \mu_t(\bm{J}_{t+1} - \bm{J}_{t}), \\
            \Rightarrow
            &\bm{0} \in \partial\left(\lambda_1 \Vert\bm{Z}^*\Vert_*\right) + \bm{M}_3^*.
        \end{aligned}
    \end{equation}
    When $t \rightarrow + \infty$, we have
    \begin{equation}
        \begin{aligned}
            \partial\mathcal{L}_{\bm{Z}}\left(   \bm{X},  \bm{Z},  \bm{B},  \bm{J},  \bm{K},  \bm{E}_f,  \bm{E}_l,  \bm{M}_{i = 1}^5 \right)|_{(\bm{Z} = \bm{Z}^*)} = 0.
        \end{aligned}
    \end{equation}
    By the updating of $\bm{B}_{t + 1}$, we can see that
    \begin{equation}
        \begin{aligned}
            \bm{0} = &-\bm{M}_{2,t+1} + \bm{M}_{4, t+1} - \mu_t(\bm{E}_{l, t+1} 
            - \bm{E}_{l, t}) + \mu_t(\bm{K}_{t+1}\bm{J}_{t+1} - \bm{K}_{t}\bm{J}_{t}).
        \end{aligned}
    \end{equation}
    With the assumption $\lim\limits_{t \rightarrow +\infty}\mu_t(\bm{E}_{l, t + 1} - \bm{E}_{l, t}) = 0$ and the inference $\lim\limits_{t \rightarrow +\infty}\mu_t(\bm{K}_{t + 1}\bm{J}_{t+1} - \bm{K}_t\bm{J}_t) = 0$,
    let $t \rightarrow + \infty$,
    and then 
    \begin{equation}
        \begin{aligned}
            \bm{0} = &-\bm{M}_2^* + \bm{M}_4^*,
        \end{aligned}
    \end{equation}
which leads to 
    \begin{equation}
        \begin{aligned}
            \partial\mathcal{L}_{\bm{B}}\left(   \bm{X},  \bm{Z},  \bm{B},  \bm{J},  \bm{K},  \bm{E}_f,  \bm{E}_l,  \bm{M}_{i = 1}^5 \right)|_{(\bm{B} = \bm{B}^*)} = 0.
        \end{aligned}
    \end{equation}

By the updating of $\bm{J}_{t+1}$, we have 
    \begin{equation}
        \begin{aligned}
            \bm{0} &= - \bm{M}_{3,t+1} - \bm{K}_t^\top\left(\bm{M}_{4,t+1} + \mu_t\left(\bm{K}_{t+1} - \bm{K}_t\right) \bm{J}_{t+1}\right)\\
            & = \left(\bm{K}_{t+1}^\top - \bm{K}_t^\top\right)\bm{M}_{4,t+1} + \bm{K}_t^\top\mu_t\left(\bm{K}_{t+1} - \bm{K}_t\right) \bm{J}_{t+1}\\
            & +\bm{M}_{3,t+1} - \bm{K}_{t+1}^\top\bm{M}_{4,t+1},
        \end{aligned}
    \end{equation}
    with $t \rightarrow +\infty$ and $\lim\limits_{t \rightarrow +\infty}\mu_t(\bm{K}_{t + 1} - \bm{K}_t) = 0$ and its inference $\lim\limits_{t \rightarrow +\infty}\bm{K}_{t + 1} - \bm{K}_t = 0$, it concludes that 

      \begin{equation}
        \begin{aligned}
            \bm{0} = \bm{M}_3^* - \bm{K}^{*\top}\bm{M}_4^*,
        \end{aligned}
    \end{equation}
    which leads to
    \begin{equation}
        \begin{aligned}
            \partial\mathcal{L}_{\bm{J}}\left(   \bm{X},  \bm{Z},  \bm{B},  \bm{J},  \bm{K},  \bm{E}_f,  \bm{E}_l,  \bm{M}_{i = 1}^5 \right)|_{(\bm{J} = \bm{J}^*)} = 0.
        \end{aligned}
    \end{equation}

    By the updating of $\bm{K}_{t+1}$,
    \begin{equation}
            \begin{aligned}
             \bm{0} = \bm{M}_{4,t + 1}\bm{J}_{t + 1}^\top + \bm{M}_{5, t + 1},
            \end{aligned}
    \end{equation}

    and then let $t \rightarrow +\infty$, we have
    \begin{equation}
            \begin{aligned}
             \bm{0} = \bm{M}_4^*\bm{J}^{*\top} + \bm{M}_5^*,
            \end{aligned}
    \end{equation}
    and then
    \begin{equation}
        \begin{aligned}
            \partial\mathcal{L}_{\bm{K}}\left(   \bm{X},  \bm{Z},  \bm{B},  \bm{J},  \bm{K},  \bm{E}_f,  \bm{E}_l,  \bm{M}_{i = 1}^5 \right)|_{(\bm{K} = \bm{K}^*)} = 0.
        \end{aligned}
    \end{equation}

    By the updating of $\bm{E}_{f, t+1}$,
    \begin{equation}
        \begin{aligned}
            \bm{0} \in \partial(\lambda_2 \Vert \bm{E}_{f, t + 1}\Vert_1) - \bm{M}_{1, t + 1},
        \end{aligned}
    \end{equation}
    
    and let $t \rightarrow +\infty$, we have

    \begin{equation}
        \begin{aligned}
            \bm{0} \in \partial(\lambda_2 \Vert \bm{E}_f^*\Vert_1) - \bm{M}_1^*,
        \end{aligned}
    \end{equation} 
    namely
    \begin{equation}
        \begin{aligned}
            \partial\mathcal{L}_{\bm{E}_f}\left(   \bm{X},  \bm{Z},  \bm{B},  \bm{J},  \bm{K},  \bm{E}_f,  \bm{E}_l,  \bm{M}_{i = 1}^5 \right)|_{(\bm{E}_f = \bm{E}_f^*)} = 0.
        \end{aligned}
    \end{equation}
    Similarly, it holds that 
    \begin{equation}
        \begin{aligned}
            \partial\mathcal{L}_{\bm{E}_l}\left(   \bm{X},  \bm{Z},  \bm{B},  \bm{J},  \bm{K},  \bm{E}_f,  \bm{E}_l,  \bm{M}_{i = 1}^5 \right)|_{(\bm{E}_l = \bm{E}_l^*)} = 0.
        \end{aligned}
    \end{equation}
    Up to now, we have proved the accumulation point $\{\Gamma^* = (\bm{X}^*, \bm{Z}^*, \bm{B}^*, \bm{J}^*, \bm{K}^*, \bm{E}_l^*, \bm{E}_f^*, \{\bm{M}_{i}^*\}_{i=1}^5 )\}$ satisfies the first-order KKT conditions of $\mathcal{L}$, and then we conclude that $\bm{X}^*,\! \bm{Z}^*,\! \bm{B}^*,\! \bm{J}^*, \!\bm{K}^*, \!\bm{E}_l^*,\! \bm{E}_f^*$ is a stationary point.
    \end{enumerate}
\end{proof}

\subsection{The Proof of Generalization Bound (Theorem 2)}
Firstly, we present some basic preliminaries.
\begin{lemma}
\label{lemma: rademacher complexity of l21}
~\cite{wei2019harnessing}
Let $\mathcal{W}$ $=\{\bm{W}:$ $\left\|\bm{W}\right\|_{2,1} \leq \mathcal{W}_{2,1} \}$ and $\mathcal{A} =\{\bm{A} \in \mathbb{R}^{n\times c}:$ $\left\|\bm{A}\right\|_{2,\infty} \leq \mathcal{A}_{2, \infty}\}$, and then the empirical Rademacher complexity of the function class with $F(\bm{W}) = \frac{1}{2}\left\|\bm{W}\right\|_{2,q}^2$ for $q = \frac{\ln\left(c\right)}{\ln\left(c\right) - 1}$ is bounded as
	\begin{equation}
		\begin{aligned}
		\mathbb{E}_\sigma[\sup_{f\in\mathcal{F}}\frac{1}{n_r}\sum_{\alpha = 1}^{n_r} \sigma_\alpha\mathrm{tr}\left(\bm{W}^\top \bm{A}^{(\alpha)}\right)] \leq \mathcal{W}_{2,1} \mathcal{A}_{2,\infty}  \sqrt{\frac{3\ln \left(c\right)}{n_r}},
		\end{aligned}
	\end{equation}
	with the fact that the dual norm of $\ell_{2,1}$ is $\ell_{2,\infty}$.
\end{lemma}

\begin{lemma}
\label{lemma: rademacher complexity of F_norm}
~\cite{wei2019harnessing}
	Let $\mathcal{W} = $ $\{\bm{W}:\left\|\bm{W}\right\|_F \leq \mathcal{W}_F \}$ and $\mathcal{A} =\{\bm{A} \in \mathbb{R}^{n\times c}: $ $\left\|\bm{A}\right\|_F \leq \mathcal{A}_F\}$, and then the empirical Rademacher complexity of the function class with $F(\bm{W}) = \frac{1}{2}\left\|\bm{W}\right\|_{2,2}^2$ is bounded as
	\begin{equation}
	\begin{aligned}
	\mathbb{E}_\sigma[\sup_{f\in\mathcal{F}}\frac{1}{n_r}\sum_{\alpha = 1}^{n_r} \sigma_\alpha\mathrm{tr}\left(\bm{W}^\top \bm{A}^{(\alpha)}\right)] \leq \mathcal{W}_F \mathcal{A}_F  \sqrt{\frac{2}{n_r}},
	\end{aligned}
	\end{equation}
	with the fact that the dual norm of Frobenius norm is Frobenius norm.
\end{lemma}

\begin{lemma}
	\label{lemma: rademacher complexity of nuclear norm}
	~\cite{chiang2015matrix} Let $\mathcal{W} = \left\{\bm{W}:\left\|\bm{W}\right\|_* \leq \mathcal{W}_* \right\}$ and $\mathcal{A} =\{\bm{A} \in \mathbb{R}^{n\times c}:$ $\left\|\bm{A}\right\|_2 \leq \mathcal{A}_2\}$, and then the empirical Rademacher complexity of the function class with $F(\bm{W}) = \frac{1}{2}\left\|\bm{W}\right\|_*^2$ is bounded as
	\begin{equation}
	\begin{aligned}
	\mathbb{E}_\sigma[\sup_{f\in\mathcal{F}}\frac{1}{n_r}\sum_{\alpha = 1}^{n_r} \sigma_\alpha\mathrm{tr}\left(\bm{W}^\top \bm{A}^{(\alpha)}\right)] \leq \mathcal{W}_* \mathcal{A}_2  \sqrt{\frac{\ln\left(2n_c\right)}{n_r}},
	\end{aligned}
	\end{equation}
	with the fact that the dual norm of the nuclear norm is spectral norm and $n_c = \max(n,c)$.
\end{lemma}

Here we give the proof of the Rademacher generalization bound theorem:
\begin{proof}
    The Rademacher complexity of our model can be written as 
\begin{equation}
\begin{aligned}
    \mathcal{R}(\mathcal{F}):= \mathbb{E}_\sigma\left[\sup\limits_{\theta \in \Theta} \frac{1}{nc} \sum\limits_{\alpha = 1}^{nc} \sigma_\alpha\left(\bm{X}_{i_\alpha}\bm{Z}\bm{I}^{j_\alpha} + \bm{E}_{l, i_\alpha, j_\alpha}\right) \right].
\end{aligned} 
\end{equation}

Since $\bm{E}_l$ is independent of $\bm{Z}$ and $\bm{X}$, the Rademacher complexity above can be rewritten as

\begin{equation}
\begin{aligned}
    \mathcal{R}(\mathcal{F})
    &:= \mathbb{E}_\sigma\left[\sup\limits_{{\bm{X},\bm{Z}} \in \Theta_{\bm{X},\bm{Z}}} \frac{1}{nc} \sum\limits_{\alpha = 1}^{nc} \sigma_\alpha\bm{X}_{i_\alpha}\bm{Z}\bm{I}^{j_\alpha} \right]
    + \mathbb{E}_\sigma\left[\sup\limits_{\bm{E} \in \Theta_{\bm{E}}} \frac{1}{nc} \sum\limits_{\alpha = 1}^{nc} \sigma_\alpha\bm{E}_{l, i_\alpha, j_\alpha}\right]\\
    &= \mathbb{E}_\sigma\left[\sup\limits_{{\bm{X},\bm{Z}} \in \Theta_{\bm{X},\bm{Z}}} \frac{1}{nc} \sum\limits_{\alpha = 1}^{nc} \sigma_\alpha \mathrm{tr}\left(\bm{Z}^\top\bm{X}^\top \bm{I}^{i_\alpha}\bm{I}^{{j_\alpha}\top}\right)\right]\\
    &+ \mathbb{E}_\sigma\left[\sup\limits_{\bm{E} \in \Theta_{\bm{E}}} \frac{1}{nc} \sum\limits_{\alpha = 1}^{nc} \sigma_\alpha\mathrm{tr}\left(\bm{E}_l^\top\bm{I}^{i_\alpha}\bm{I}^{{j_\alpha}\top}\right)\right]\\
    &= \mathbb{E}_\sigma\left[\sup\limits_{{\bm{X},\bm{Z}} \in \Theta_{\bm{X},\bm{Z}}} \frac{1}{nc} \sum\limits_{\alpha = 1}^{nc} \sigma_\alpha \langle\bm{X}\bm{Z}, \bm{I}^{i_\alpha}\bm{I}^{{j_\alpha}\top}\rangle\right]\\
    &+ \mathbb{E}_\sigma\left[\sup\limits_{\bm{E} \in \Theta_{\bm{E}}} \frac{1}{nc} \sum\limits_{\alpha = 1}^{nc} \sigma_\alpha \langle\bm{E}_l, \bm{I}^{i_\alpha}\bm{I}^{{j_\alpha}\top}\rangle \right],
\end{aligned} 
\end{equation}
where $\Theta_{\bm{E}} = \{ \bm{E}|\Vert\bm{E}_l\Vert_{2,1} \leq \mathcal{E}_{l,21}\}$, and $\Theta_{\bm{X},\bm{Z}} = \{(\bm{X}, \bm{Z})|\Vert\bm{Z}\Vert_* \leq \mathcal{Z}_*, \Vert\bm{X}\Vert_* \leq \mathcal{X}_*, \Vert\bm{\tilde{X}} - \bm{X}\Vert_1 \leq \mathcal{E}_{f,1}, \bm{XZ} \in [0,1]^{n \times c}\}$.

As $\bm{X} = \bm{\tilde{X}} - \bm{E}_f$, and thus 

\begin{equation}
    \begin{aligned}
    \label{eq: Rademacher complexity with several contraints}
    \mathcal{R}(\mathcal{F})
    & \leq 
    \min\left\{\mathbb{E}_\sigma\left[\sup\limits_{{\bm{X}_1,\bm{Z}} \in \Theta_{\bm{X}_1,\bm{Z}}} \frac{1}{nc} \sum\limits_{\alpha = 1}^{nc} \sigma_\alpha \langle\bm{X}\bm{Z}, \bm{I}^{i_\alpha}\bm{I}^{{j_\alpha}\top}\rangle\right],\right.\\
     &\ \ \ \left.\mathbb{E}_\sigma\left[\sup\limits_{{\bm{E}_f,\bm{Z}} \in \Theta_{\bm{E}_f,\bm{Z}}} \frac{1}{nc} \sum\limits_{\alpha = 1}^{nc} \sigma_\alpha \langle\bm{Z}, (\bm{\tilde{X}} - \bm{E}_f)^\top \bm{I}^{i_\alpha}\bm{I}^{{j_\alpha}\top}\rangle\right]\right\}\\
    &\ \ \ + \mathbb{E}_\sigma\left[\sup\limits_{\bm{E} \in \Theta_{\bm{E}}} \frac{1}{nc} \sum\limits_{\alpha = 1}^{nc} \sigma_\alpha \langle\bm{E}_l, \bm{I}^{i_\alpha}\bm{I}^{{j_\alpha}\top}\rangle \right],
    \end{aligned}
\end{equation}
where $\Theta_{\bm{X}_1,\bm{Z}} = \{(\bm{X}, \bm{Z})|\Vert\bm{XZ}\Vert_* \leq  \mathcal{X}_*\mathcal{Z}_*, \Vert\bm{XZ}\Vert_F \leq \sqrt{n} \}$, and $\Theta_{\bm{E}_f,\bm{Z}} = \{(\bm{E}_f, \bm{Z})|\Vert\bm{Z}\Vert_* \leq \mathcal{Z}_*, \Vert\bm{E}_f\Vert \leq \mathcal{E}_{f,1}\}$.

As $\Vert\bm{\tilde{X}} - \bm{E}_f\Vert_F \leq \Vert\bm{\tilde{X}} \Vert_F + \Vert\bm{E}_f\Vert_F \leq \Vert\bm{\tilde{X}} \Vert_F + \sqrt{d}\Vert\bm{E}_f\Vert_1$, it holds that

\begin{equation}
    \begin{aligned}
        \mathbb{E}_\sigma&\left[\sup\limits_{{\bm{E}_f,\bm{Z}} \in \Theta_{\bm{E}_f,\bm{Z}}} \frac{1}{nc} \sum\limits_{\alpha = 1}^{nc} \sigma_\alpha \langle\bm{Z}, (\bm{\tilde{X}} - \bm{E}_f)^\top \bm{I}^{i_\alpha}\bm{I}^{{j_\alpha}\top}\rangle\right]\\
      = & \frac{1}{nc}\mathbb{E}_\sigma\left[\sup\limits_{{\bm{E}_f,\bm{Z}} \in \Theta_{\bm{E}_f,\bm{Z}}} \langle\bm{Z}, (\bm{\tilde{X}} - \bm{E}_f)^\top \left(\sum\limits_{\alpha = 1}^{nc} \sigma_\alpha\bm{I}^{i_\alpha}\bm{I}^{{j_\alpha}\top}\right)\rangle\right]\\
      \leq & \frac{1}{nc}\mathbb{E}_\sigma\sup\limits_{{\bm{E}_f,\bm{Z}} \in \Theta_{\bm{E}_f,\bm{Z}}} \Vert\bm{Z}\Vert_*\Vert(\bm{\tilde{X}} - \bm{E}_f)^\top \left(\sum\limits_{\alpha = 1}^{nc} \sigma_\alpha\bm{I}^{i_\alpha}\bm{I}^{{j_\alpha}\top}\right)\Vert_F\\
      \leq & \frac{1}{nc}\mathbb{E}_\sigma\sup\limits_{{\bm{E}_f,\bm{Z}} \in \Theta_{\bm{E}_f,\bm{Z}}} \Vert\bm{Z}\Vert_* \Vert\bm{\tilde{X}} - \bm{E}_f\Vert_F \Vert\sum\limits_{\alpha = 1}^{nc} \sigma_\alpha\bm{I}^{i_\alpha}\bm{I}^{{j_\alpha}\top}\Vert_F\\
      \leq & \frac{1}{nc} \mathcal{Z}_*(\mathcal{\tilde{X}}_F + \sqrt{d}\mathcal{E}_{f,1})\sqrt{nc} = \frac{1}{\sqrt{nc}}\mathcal{Z}_*(\mathcal{\tilde{X}}_F + \sqrt{d}\mathcal{E}_{f,1}).
    \end{aligned}
\end{equation}

Taking Lemma~\ref{lemma: rademacher complexity of l21},~\ref{lemma: rademacher complexity of F_norm}, and~\ref{lemma: rademacher complexity of nuclear norm} into consideration, Eq.~\eqref{eq: Rademacher complexity with several contraints} leads to
\begin{equation}
    \begin{aligned}
    \mathcal{R}(\mathcal{F})
    & \leq 
    \mathcal{E}_{l,21}\Vert\bm{I}^{i}\bm{I}^{{j}\top}\Vert_{2, \infty}\sqrt{\frac{3\ln{c}}{nc}} \\
    &+ \min\left\{\mathcal{X}_*\mathcal{Z}_*\Vert\bm{I}^{i}\bm{I}^{{j}\top}\Vert_2\sqrt{\frac{\ln{2n_c}}{nc}}, \frac{1}{\sqrt{nc}}\mathcal{Z}_*(\mathcal{\tilde{X}}_F + \sqrt{d}\mathcal{E}_{f,1}), \sqrt{n}\Vert\bm{I}^{i}\bm{I}^{{j}\top}\Vert_F\sqrt{\frac{2}{nc}}\right\}.
    \end{aligned}
\end{equation}
Since $\max\limits_{i,j}\Vert\bm{I}^{i}\bm{I}^{{j}\top}\Vert_{2, \infty} = 1$, $\max\limits_{i,j}\Vert\bm{I}^{i}\bm{I}^{{j}\top}\Vert_2 = 1$, and $\max\limits_{i,j}\Vert\bm{I}^{i}\bm{I}^{{j}\top}\Vert_F = 1$, we arrive at the upper bound of $\mathcal{R}_n(\mathcal{F}_\Theta)$, which is 

\begin{equation}
    \begin{aligned}
    \mathcal{R}_n(\mathcal{F}_\Theta)
    & \leq 
    \mathcal{E}_{l,21}\sqrt{\frac{3\ln{c}}{nc}} + 
     \min\left\{\mathcal{X}_*\mathcal{Z}_*\sqrt{\frac{\ln{(2n_c)}}{nc}}, \frac{1}{\sqrt{nc}}\mathcal{Z}_*(\mathcal{\tilde{X}}_F + \sqrt{d}\mathcal{E}_{f,1}), \sqrt{\frac{2}{c}}\right\}.
    \end{aligned}
\end{equation}
\end{proof}
\end{document}